\documentclass[runningheads]{llncs}
\usepackage{graphicx}
\usepackage{comment}
\usepackage{amsmath,amssymb,bm} % define this before the line numbering.
\usepackage{color}

\usepackage[utf8]{inputenc}
\usepackage{graphicx}
\usepackage{epsfig}
\usepackage{caption}
\usepackage{subcaption}
\usepackage{multirow}
\usepackage{array}
\usepackage{arydshln}
\usepackage[ruled,vlined]{algorithm2e}
\usepackage{algorithmic,float}
\usepackage{rotating}
\usepackage{chngcntr}

\usepackage{dsfont}
\usepackage{bbm}
\usepackage{times}
\usepackage[resetlabels,labeled]{multibib}
\newcommand{\etal}{\textit{et al}. }

\DeclareMathOperator*{\argmin}{arg\,min}
\def\A{{\mathcal{A}}}

\def\rank{{\text{ rank}}}
\def\diag{{\text{ diag}}}

\def\bfsigma{{\bm \sigma}}
\def\bfgamma{{\bm \gamma}}

\renewcommand{\vec}[1]{\text{vec}\left( #1 \right)}

\begin{document}
% \linenumbers
\pagestyle{headings}
\mainmatter
\def\ECCVSubNumber{5592}  % Insert your submission number here

\title{Accurate Optimization of Weighted Nuclear Norm for Non-Rigid Structure from Motion} % Replace with your title

\begin{comment}
\titlerunning{ECCV-20 submission ID \ECCVSubNumber} 
\authorrunning{ECCV-20 submission ID \ECCVSubNumber} 
\author{Anonymous ECCV submission}
\institute{Paper ID \ECCVSubNumber}
\end{comment}
%******************

%\begin{comment}
\titlerunning{Accurate Optimization of Weighted Nuclear Norm for NRSfM}
\author{José Pedro Iglesias\inst{1} \and
Carl Olsson\inst{1,2} \and
Marcus Valtonen Örnhag \inst{2}}
\authorrunning{J.P. Iglesias et al.}
\institute{Chalmers University of Technology, Sweden \and
Lund University, Sweden }
%\end{comment}
%******************
\maketitle

\begin{abstract}
Fitting a matrix of a given rank to data in a least squares sense can be done very effectively using 2nd order methods such as Levenberg-Marquardt by explicitly optimizing over a bilinear parameterization of the matrix. In contrast, when applying more general singular value penalties, such as weighted nuclear norm priors, direct optimization over the elements of the matrix is typically used. Due to non-differentiability of the resulting objective function, first order sub-gradient or splitting methods are predominantly used. While these offer rapid iterations it is well known that they become inefficent near the minimum due to zig-zagging and in practice one is therefore often forced to settle for an approximate solution.

In this paper we show that more accurate results can in many cases be achieved with 2nd order methods. Our main result shows how to construct bilinear formulations, for a general class of regularizers including weighted nuclear norm penalties, that are provably equivalent to the original problems. With these formulations the regularizing function becomes twice differentiable and 2nd order methods can be applied. We show experimentally, on a number of structure from motion problems, that our approach outperforms state-of-the-art methods.\footnote[1,2]{This work was supported by the Swedish Research Council (grants no. 2015-05639, 2016-04445 and 2018-05375), the Swedish Foundation for Strategic Research
(Semantic Mapping and Visual Navigation for Smart Robots) and the Wallenberg AI, Autonomous Systems and Software Program (WASP) funded by the Knut and Alice Wallenberg Foundation.}
%\keywords{We would like to encourage you to list your keywords within the abstract section}
\end{abstract}

%===========================================================
\section{Introduction}
Matrix recovery problems of the form
\begin{equation}
\min_{X} f(\bfsigma(X))) + \|\A X - b\|^2,
\label{eq:general}
\end{equation}
where $\A$ is a linear operator and $\bfsigma(X) = (\sigma_1(X),\sigma_2(X),...)$ are the singular values of $X$,
are frequently occurring in computer vision. Applications range from high level 3D reconstruction problems to low level pixel manipulations \cite{tomasi-kanade-ijcv-1992,bregler-etal-cvpr-2000,yan-pollefeys-pami-2008,garg-etal-cvpr-2013,basri-etal-ijcv-2007,garg-etal-ijcv-2013,wang-etal-2012,canyi-etal-cvpr-2014,oh-etal-pami-2016,hu-etal-pami-2013,gu-2016}.
In structure from motion the most common approaches enforce a given low rank $r$ without additionally penalizing non-zero singular values \cite{tomasi-kanade-ijcv-1992,bregler-etal-cvpr-2000,hong-fitzgibbon-cvpr-2015}. 
(This can be viewed as a special case of \eqref{eq:general} by letting $f$ assign zero if fewer than $r$ singular values are non-zero and infinity otherwise.)

Since the rank of a matrix $X$ is bounded by to the number of columns/rows in a bilinear parameterization $X = B C^T$, the resulting optimization problem can be written
\begin{equation}
\min_{B,C} \|\A(BC^T)-b\|^2.
\end{equation}
This gives a smooth objective function and can therefore be optimized using 2nd order methods.
In structure from motion problems, where the main interest is the extraction of camera matrices from $B$ and 3D points from $C$, this is typically the preferred option \cite{buchanan-fitzgibbon-cvpr-2005}. In a series of recent papers Hong \etal showed that optimization with the VarPro algorithm is remarkably robust to local minima converging to accurate solutions  \cite{hong-fitzgibbon-cvpr-2015,hong-etal-eccv-2016,hong-etal-cvpr-2017}. 
In \cite{hong-zach-cvpr-2018} they further showed how uncalibrated rigid structure from motion with a proper perspective projection can be solved within a factorization framework. 
On the downside the accuracy of these methods comes with a price.
Typically the iterations are costly since (even when the Schur complement trick is used) 2nd order methods require an inversion of a relatively large hessian matrix, which may hinder application when suitable sparsity patterns are not present.

For low level vision problems such as denoising and inpainting, eg.
\cite{oh-etal-pami-2016,hu-etal-pami-2013,gu-2016}, the main interest is to recover the elements of $X$ and not the factorization. In this context more general regularization terms that also consider the size of the singular values are often used. Since the singular values are non-differentiable functions of the elements in $X$ first order methods are usually employed. The simplest option is perhaps a splitting methods such as ADMM \cite{boyd-etal-2011} since the proximal operator
\begin{equation}
\argmin_X f(\sigma(X)) + \|X-X_0\|^2,
\end{equation}
can often be computed in closed form \cite{hu-etal-pami-2013,gu-2016,oh-etal-pami-2016,dai-etal-ijcv-2014,kumar-arxiv-2019}. Alternatively, subgradient methods can be used to handle the non-differentiability of the regularization term \cite{canyi-etal-cvpr-2014}.

It is well known that while first order methods have rapid iterations and make large improvements the first couple of iterations they have a tendency to converge slowly when approaching the optimum.
For example, \cite{boyd-etal-2011} recommends to use ADMM when a solution in the vicinity of the optimal point is acceptable, but suggests to switch to a higher order method when high accuracy is desired. 
For low level vision problems where success is not dependent on achieving an exact factorization of a particular size, first order methods may therefore be suitable. In contrast, in the context of structure from motion, having roughly estimated elements in $X$ causes the obtained factorization $B$, $C$ to be of a much larger size than necessary yielding poor reconstructions with too many deformation modes.

In this paper we aim to extend the class of methods that can be optimized using bilinear parameterization allowing accurate estimation of a low rank factorization from a general class of regularization terms. While our theory is applicable for many objectives we focus on weighted nuclear norm penalties since these have been successfully used in structure form motion applications. We show that these can be optimized with 2nd order methods which significantly increases the accuracy of the reconstruction. We further show that with these improvements the model of Hong \etal \cite{hong-zach-cvpr-2018} can be extended to handle non-rigid reconstruction with a proper perspective model, as opposed to the orthographic projection model adopted by other factorization based approaches, e.g. \cite{kumar-arxiv-2019,dai-etal-ijcv-2014,garg-etal-cvpr-2013,yan-pollefeys-pami-2008}.

\subsection{Related Work and Contributions}

Minimization directly over $X$ has been made popular since the problem is convex when $f$ is convex and absolutely symmetric, that is, $f(|x|) = f(x)$ and $f(\Pi x) = f(x)$, where $\Pi$ is any permutation \cite{lewis1995convex}.
Convex penalties are however of limited interest since they generally prefer solutions with many small non-zero singular values to those with few large ones. A notable exception is the nuclear norm  \cite{fazel-etal-acc-2015,recht-etal-siam-2010,oymak2011simplified,candes-etal-acm-2011,candes2009exact} which penalizes the sum of the singular values. Under the RIP assumption \cite{recht-etal-siam-2010} exact or approximate low rank matrix recovery can then be guaranteed \cite{recht-etal-siam-2010,candes2009exact}.
On the other hand, since the nuclear norm penalizes large singular values, it suffers from a shrinking bias \cite{cabral-etal-iccv-2013,canyi-etal-cvpr-2014,larsson-olsson-ijcv-2016}.

An alternative approach that unifies bilinear parameterization with regularization approaches is based on the observation \cite{recht-etal-siam-2010} that the nuclear norm $\|X\|_*$ of a matrix $X$ can be expressed as
$
\|X\|_* = \min_{BC^T = X} \frac{\|B\|_F^2+\|C\|_F^2}{2}.
$
Thus when $f(\bfsigma(X)) = \mu \sum_i\sigma_i(X)$, where $\mu$ is a scalar controlling the strength of the regularization, optimization of \eqref{eq:general} can be formulated as 
{\small
\begin{equation}
\min_{B,C} \mu \frac{\|B\|_F^2+\|C\|_F^2}{2}+ \|\A BC^T - b\|^2.
\label{eq:nuclearbilin}
\end{equation}}\normalsize
Optimizing directly over the factors has the advantages that the number of variables is much smaller and the objective function is two times differentiable so second order methods can be employed. 
While \eqref{eq:nuclearbilin} is non-convex because of the bilinear terms, the convexity of the nuclear norm can still be used to show that any local minimizer $B$,$C$ with $\rank(B C^T) < k$, where $k$ is the number of columns in $B$ and $C$, is globally optimal \cite{bach-arxiv-2013,haeffele-vidal-arxiv-2017}. 
The formulation \eqref{eq:nuclearbilin} was for vision problems in \cite{cabral-etal-iccv-2013}. In practice it was observed that the shrinking bias of the nuclear norm makes it too weak to enforce a low rank when the data is noisy. Therefore, a ``continuation'' approach where the size of the factorization is gradually reduced was proposed. While this yields solutions with lower rank, the optimality guarantees no longer apply. 
Bach \etal \cite{bach-arxiv-2013} showed that
{\small
\begin{equation}
\|X\|_{s,t}:=\min_{X=BC^T} \sum_{i=1}^k\frac{\|B_i\|_s^2 + \|C_i\|_t^2}{2},
\label{eq:decomposition-norm}
\end{equation}}\normalsize
where $B_i$,$C_i$ are the $i$th columns of $B$ and $C$ respectively,
is convex for any choice of vector norms $\|\cdot\|_s$ and \mbox{$\|\cdot\|_t$}.
In \cite{haeffele-vidal-arxiv-2017} it was shown that a more general class of 2-homogeneous factor penalties result in a convex regularization similar to \eqref{eq:decomposition-norm}. The property that a local minimizer $B$, $C$ with $\rank(B C^T) < k$, is global is also extended to this case. Still, because of convexity, it is clear that these formulations will suffer from a similar shrinking bias as the nuclear norm.

One way of reducing shrinking bias is to use penalties that are constant for large singular values.
Shang \etal \cite{shang-etal-2018} showed that penalization with the Schatten semi-norms $\|X\|_q = \sqrt[q]{\sum_{i=1}^N \sigma_i(X)^q}$, for $q=1/2$ and $2/3$, can be achieved using a convex penalty on the factors $B$ and $C$. A generalization to general values of $q$ is given in \cite{xu-etal-AAAI-2017}.
An algorithm that address a general class of penalties for symmetric matrices is presented in \cite{krechetov-etal-2019}.
In \cite{valtonen-ornhag-etal-2018-arxiv} it was shown that if $f$ is separable with the same penalty for each singular value, that is, $f(\bfsigma(X)) = \sum_i g(\sigma_i(X))$, where $g$ is differentiable, concave and non-decreasing then \eqref{eq:general} can be optimized using second order methods such as Levenberg-Marquart or VarPro. This is achieved by re-parameterizing the matrix $X$ using a bilinear factorization $X = B C^T$ and optimizing
\begin{equation}
\min_{B,C} f(\bfgamma(B,C))+\|\A(BC^T) + b\|^2.
\label{eq:generalbilinear}
\end{equation} 
Here $\bfgamma(B,C) = (\gamma_1(B,C),\gamma_2(B,C),...)$ and $\gamma_i(B,C) = \frac{\|B_i\|^2+\|C_i\|^2}{2}$. In contrast to the singular value $\sigma_i(X)$ the function 
$\gamma_i(B,C)$ is smooth which allows optimization with second order methods.
It is shown in \cite{valtonen-ornhag-etal-2018-arxiv} that if $X^*$ is optimal in \eqref{eq:general} 
then the factorization $B=L\sqrt{\Sigma}, C = R\sqrt{\Sigma}$, where $X^* = L \Sigma R^T$ is the SVD of $X^*$, is optimal in \eqref{eq:generalbilinear}. (Here we assume that $L$ is $m \times r$, $\Sigma$ is $r \times r$ and $R$ is $n\times r$, with $\rank(X)=r$.) Note also that this choice gives $\gamma_i(B,C) = \sigma_i(X^*)$.

Having a separable objective with the same penalty for each singular value is however somewhat restrictive. 
An alternative way of reducing bias is to re-weight the nuclear norm and use $f(\bfsigma(X)) = \sum_i a_i \sigma_i(X)$ \cite{hu-etal-pami-2013,gu-2016,kumar-arxiv-2019}. Assigning low weights to the first (largest) singular values allows accurate matrix recovery. In addition the weights can be used to regularize the size of the non-zero singular values which has been shown to be an additional useful prior in NRSfM \cite{kumar-arxiv-2019}. Note however that the singular values are always ordered in non-increasing order. Therefore, while the function is linear in the singular values it is in fact non-convex and non-differentiable in the elements of $X$ whenever the singular values are not distinct (typically the case in low rank recovery). 

In this paper we show that this type of penalties allow optimization with $\bfgamma(B,C)$ instead of $\bfsigma(X)$. 
In particular we study the optimization problem
{\small
\begin{eqnarray}
& \min_{B,C} & f(\bfgamma(B,C)) \label{eq:regobj}\\
& \text{s.t.} & BC^T = X, \label{eq:regconst}
\end{eqnarray}}
and its constraint set for a fixed $X$. We characterize the extreme-points of the feasible set using permutation matrices and give conditions on $f$ that ensure that the optimal solution is of the form $\bfgamma(B,C) = \Pi \bfsigma(X)$, where $\Pi$ is a permutation.
For the weighted nuclear norm $f(\bfsigma(X)) = a^T \bfsigma(X)$ we show that if the elements of $a$ are non-decreasing the optimal solution has $\bfgamma(B^*,C^*) = \bfsigma(X)$. A simple consequence of this result is that 
\begin{equation}
 \min_{B,C} a^T \bfgamma(B,C)) + \|\A (B C^T) - b\|^2
 \label{eq:wnnbilinear}
\end{equation}
is equivalent to
\begin{equation}
\min_X a^T \bfsigma(X) + \|\A X - b\|^2.
\end{equation}
While the latter is non-differentiable the former is smooth and can be minimized efficiently with second order methods. 

Our experimental evaluation confirms that this approach outperforms current first order methods in terms of accuracy as can be expected. On the other hand first order methods make large improvments the first coupler of iterations and therefore we combine the two approaches. We start out with a simple ADMM implementation and switch to our second order approach when only minor progress is being made. Note however that since the original formulation is non-convex local minima can exist. In addition bilinear parameterization introduces additional stationary points that are not present in the original $X$ parameterization. One such example is $(B,C)=(0,0)$, where all gradients vanish. Still our experiments show that the combination of these methods often converge to a good solution from random initialization.

\section{Bilinear Parameterization Penalties}
In this section we will derive a dependence between the singular values $\sigma_i(X)$ and the $\gamma_i(B,C)$, when $BC^T = X$.
For ease of notation we will suppress the dependence on $X$ and $(B,C)$ since this will be clear from the context.
Let $X$ have the SVD $X = R\Sigma L^T$, $B=R\sqrt{\Sigma}$ and $C = L\sqrt{\Sigma}$. 
We will study other potential factorizations $X = \hat{B}\hat{C}^T$ using 
$\hat{B} = BV$, $\hat{C} = CH$ and $VH^T = I_{r \times r}$. 
In this section we will further assume that $V$ is a square $r \times r$ matrix and therefore $H^T$ is its inverse.
(We will generalize the results to the rectangular case in Section~\ref{sec:generalmat}).

We begin by noting that $\gamma_j = \frac{\|\hat{B}_j\|^2+\|\hat{C}_j\|^2}{2} = \frac{\|BV_j\|^2+\|CH_j\|^2}{2}$, where $V_j$ and $H_j$ are columns $j$ of $V$ and $H$ respectively. 
We have $\|BV_j\|^2 = V_j^T B^T B V_j = V_j^T \Sigma V_j = \|\sqrt{\Sigma} V_j\|^2,$
%{\small
%\begin{equation}
%\|BV_j\|^2 = V_j^T B^T B V_j = V_j^T \Sigma V_j = \|\sqrt{\Sigma} V_j\|^2,
%\end{equation}}
and similarly $\|CH_j\|^2=\|\sqrt{\Sigma} H_j\|^2$ and therefore $\gamma_j = \frac{\|\sqrt{\Sigma}V_j\|^2+\|\sqrt{\Sigma}H_j\|^2}{2}$.
This gives
{\small
\begin{equation}
\gamma_j = \left(\frac{\sigma_1(v_{1j}^2+h_{1j}^2)+\sigma_2(v_{2j}^2+h_{2j}^2)+...+\sigma_r(v_{rj}^2+h_{rj}^2)}{2}\right),
\end{equation}}
or in matrix form
{\small
\begin{equation}
\left(
\begin{matrix}
\gamma_1 \\
\gamma_2 \\
\vdots \\
\gamma_r
\end{matrix}
\right) = 
\frac{1}{2}\underbrace{\left(
\begin{matrix}
v_{11}^2 & v_{21}^2 & \hdots & v_{r1}^2 \\
v_{12}^2 & v_{22}^2 & \hdots & v_{r2}^2 \\
\vdots & \vdots & \ddots & \vdots \\
v_{1r}^2 & v_{2r}^2 & \hdots & v_{rr}^2 \\
\end{matrix}
\right)}_{=V^T \odot V^T}
\left(
\begin{matrix}
\sigma_1 \\
\sigma_2 \\
\vdots \\
\sigma_r
\end{matrix}
\right)
+
\frac{1}{2}\underbrace{\left(
\begin{matrix}
h_{11}^2 & h_{21}^2 & \hdots & h_{r1}^2 \\
h_{12}^2 & h_{22}^2 & \hdots & h_{r2}^2 \\
\vdots & \vdots & \ddots & \vdots \\
h_{1r}^2 & h_{2r}^2 & \hdots & h_{rr}^2 \\
\end{matrix}
\right)}_{=H^T\odot H^T}
\left(
\begin{matrix}
\sigma_1 \\
\sigma_2 \\
\vdots \\
\sigma_r
\end{matrix}
\right).
\label{eq:matrixform}
\end{equation}}
Minimizing \eqref{eq:regobj} over different factorizations is therefore equivalent to solving
\begin{eqnarray}
&\min_{\bfgamma, M \in \mathcal{S}} & f(\bfgamma), \label{eq:Sobj} \\
& \text{s.t.} & \bfgamma = M \bfsigma. \label{eq:Sconst}
\end{eqnarray}
where
{\small
\begin{equation}
\mathcal{S} = \{\frac{1}{2}(V^T\odot V^T + H^T \odot H^T); \ VH^T = I\}.
\end{equation}}
\iffalse
{\small
\begin{eqnarray}
&\min_{\bfgamma,V,H}& f(\bfgamma), \label{eq:optprobl} \\
& & \bfgamma = \frac{1}{2}(V^T \odot V^T + H^T \odot H^T) \bfsigma \\
& & VH^T = H^T V = I \label{eq:optconst}.
\end{eqnarray}}
\fi
It is clear that $V = H = \Pi^T$, where $\Pi$ is any permutation, is feasible in the above problem since permutations are orthogonal. In addition they contain only zeros and ones and therefore it is easy to see that this choice gives $\bfgamma = \frac{1}{2}(\Pi \odot \Pi + \Pi \odot \Pi) \bfsigma = \Pi \bfsigma$. 
In the next section we will show that these are extreme points of the feasible set, in the sense that they can not be written as convex combinations of other points in the set. 
Extreme points are important for optimization since the global minimum is guaranteed to be attained (if it exists) in such a point if the objective function has concavity properties.
This is for example true if $f$ is quasi-concave, that is, the super-level sets $S_{\alpha}=\{x\in \mathbb{R}^r_{\geq 0}; f(x)\geq \alpha\}$ are convex. To see this let $x = \lambda x_1+(1-\lambda)x_2$, and consider the super-level set $S_{\alpha}$ where $\alpha = \min(f(x_1),f(x_2))$. Since both $x_1 \in S_{\alpha}$ and $x_2 \in S_{\alpha}$ it is clear by convexity that so is $x$ and therefore $f(x) \geq \min(f(x_1),f(x_2))$. 

\subsection{Extreme Points and Optimality}\label{sec:extremepts}
We now consider the optimization problem \eqref{eq:Sobj}-\eqref{eq:Sconst} 
and a convex relaxation of the constraint set. For this purpose we let $\mathcal{D}$ be the set of doubly stochastic matrices 
{\small
\begin{equation}
\mathcal{D} = \{M\in \mathbb{R}^{r \times r}; \ m_{ij} \geq 0, \ \sum_i m_{ij} = 1, \ \sum_j m_{ij} = 1\}.
\end{equation}}
Note that if $V$ is orthogonal, and therefore $H=V$, then the row sum
$\sum_{j = 1}^r\frac{v_{ij}^2 + h_{ij}^2}{2}$, and the column sum  $\sum_{i = 1}^r\frac{v_{ij}^2 + h_{ij}^2}{2}$ are both one. Hence such a matrix is in $\mathcal{D}$.
To handle non-orthogonal matrices we define the set of superstochastic matrices $\mathcal{S}_W$ as all matrices $M=D+N$, where $D\in \mathcal{D}$ and $N$ is a matrix with non-negative elements. 

It can be shown that (see Theorem~6 in~\cite{bhatia-jain-2015}) that $\mathcal{S} \subset \mathcal{S}_W$. In addition it is easy to see that $\mathcal{S}_W$ is convex since it consists of affine constraints.
Therefore the problem 
\begin{eqnarray}
&\min_{\bfgamma, M \in \mathcal{S}_W} & f(\bfgamma), \label{eq:Cobj}\\
& \text{s.t.} & \bfgamma = M \bfsigma. \label{eq:Cconst}
\end{eqnarray}
is a relaxation of \eqref{eq:Sobj}-\eqref{eq:Sconst}. 
We will next show that the two problems have the same minimum if a minimizer to \eqref{eq:Cobj}-\eqref{eq:Cconst} exists when the objective function $f$ is quasi-concave (on $\mathbb{R}^r_{\geq 0}$). 
As we have seen previously, the minimum (over $\mathcal{S}_W$) is then attained in an extreme point of $\mathcal{S}_W$. 
We therefore need the following characterization. 
\begin{lemma}
	The extreme points of $\mathcal{S}_W$ are $r \times r$ permutation matrices.
\end{lemma}
\begin{proof}
First we note that any extreme point of $\mathcal{S}_W$ has to be in $\mathcal{D}$ since if $M=D+N$ with $N \neq 0$ then $M=\frac{1}{2}D+\frac{1}{2}(D+2N)$, which is a convex combination of two points in $\mathcal{S}_W$. 
By Birkhoff's Theorem \cite{birkhoff-1946} any matrix in $\mathcal{D}$ can be written as a convex combination of permutation matrices. 
\end{proof}

Since permutation matrices are orthogonal with $0/1$ elements it is clear they can be written
$\Pi = \frac{1}{2}(\Pi \odot \Pi + \Pi \odot \Pi), $
with $\Pi \Pi^T = I$. Therefore the extreme points of $\mathcal{S}_W$ are also in $\mathcal{S}$. 
Hence if the minimum of \eqref{eq:Cobj}-\eqref{eq:Cconst} is attained, there is an optimal extreme point of $\mathcal{S}_W$ which also solves \eqref{eq:Sobj}-\eqref{eq:Sconst}, and therefore the solution is given by a permutation $V=H=\Pi$. 

We conclude this section by giving sufficient conditions for the minimum of \eqref{eq:Cobj}-\eqref{eq:Cconst} to exist, namely that $f$ is lower semi-continuous and  non-decreasing in all of its variables, that is, 
if $\tilde{\gamma}_i \geq \gamma_i$ for all $i$ then $f(\tilde{\bfgamma}) \geq f(\bfgamma)$.
Since the singular values are all positive it is clear that the elements of $(D+N)\bfsigma$ are larger than those of $D\bfsigma$. Hence when $f$ is non-decreasing it is enough to consider minimization over $\mathcal{D}$. We then have a lower semi-continuous objective function on a compact set for which the minimum is known to be attained.

We can now summarize the results of this section in the following theorem:
\begin{theorem}
	Let $f$ be quasi-concave (and lower semi-continuous) on $\mathbb{R}^r_{\geq 0}$ fulfilling $f(\tilde{\bfgamma}) \geq f(\bfgamma)$ when
	$\tilde{\gamma}_i \geq \gamma_i$ for all $i$. Then there is an optimal $\bfgamma^*$ of \eqref{eq:Sobj}-\eqref{eq:Sconst} that is of the form  $\bfgamma^* = \Pi \bfsigma$ where $\Pi$ is a permutation.
\end{theorem}

\section{Non-square Matrices}\label{sec:generalmat}
In the previous section we made the assumption that $V$ and $H$ where square matrices, which corresponds to  searching over $\hat{B}$ and $\hat{C}$ consisting of $r$ columns when $\rank(X) = r$. In addition since $V$ and $H$ are invertible this means that $\hat{B}$ and $\hat{C}$ have linearly independent columns.
In this section we generalize the result from Section~\ref{sec:extremepts} to rectangular matrices $V$ and $H$. Therefore we suppose that $V$ and $H$ are non-square of size $r \times p$, $p > r$, with $VH^T = I_{r \times r}$,
and consider the slightly modified problem
{\small
\begin{eqnarray}
&\min_{\bfgamma,V,H}& f(\bfgamma), \label{eq:optproblrec} \\
& & \bfgamma = \frac{1}{2}(V^T \odot V^T + H^T \odot H^T) \bfsigma \\
& & VH^T = I_{r \times r} \label{eq:optconstrec}
\end{eqnarray}}
Note that $VH^T$ do not commute and we therefore only assume that $V$ is a left inverse of $H^T$.
In what follows we show that by adding zeros to the vector $\sigma$ we can extend $V$, $H$ into square matrices without changing the objective function.

Note that we may assume that $V$ has full row rank since otherwise $X \neq BVH^TC^T$.
Let $V^\dagger$ be the Moore-Penrose pseudo inverse and $O_{V_\perp}$ a $(p-r) \times p$ matrix containing a basis for the space orthogonal to the row space of $V$ (and the column space of $V^\dagger$).
Since $VH^T = I_{r \times r}$ the matrix $H^T$ is of the form $H^T = V^\dagger +  O^T_{V_\perp}K_1$, where $K_1$ is a $(p-r)\times r$ coefficient matrix.
We now want to find matrices $\tilde{V}$ and $\tilde{H}$ such that
{\small
\begin{equation}
\begin{bmatrix}
V \\
\tilde{V}
\end{bmatrix}
\begin{bmatrix}
V^\dagger +  O^T_{V_\perp}K_1 & \tilde{H}^T
\end{bmatrix}
=
\begin{bmatrix}
I_{r\times r} & 0 \\
0 & I_{(p-r)\times(p-r)}
\end{bmatrix}.
\end{equation}}
To do this we first select $\tilde{H}^T = O_{V_{\perp}}^T$ since $V O_{V_{\perp}}^T = 0$.
Then we let $\tilde{V} = O_{V_\perp} + K_2 V$, where $K_2$ is a size $(p-r)\times r$ coefficient matrix, since this gives $\tilde{V} \tilde{H}^T = I_{(p-r)\times(p-r)}$.
To determine $K_2$ we consider $\tilde{V}(V^\dagger +  O^T_{V_\perp}K_1) = K_2 I_{r \times r} + I_{(p-r)\times(p-r)}K_1 = K_2+K_1$. Selecting $K_2 = -K_1$ thus gives square matrices such that
{\small
\begin{equation}
\begin{bmatrix}
V \\
\tilde{V}
\end{bmatrix}
\begin{bmatrix}
H^T \tilde{H}^T
\end{bmatrix}
= I.
\end{equation}}
Further letting $\tilde{\Sigma} = \begin{bmatrix}
\Sigma & 0 \\
0 & 0
\end{bmatrix}$ shows that $\|B V_i\| = \|\sqrt{\tilde{\Sigma}} \begin{bmatrix}
V_i \\
\tilde{V}_i
\end{bmatrix}\|$ and $\|C H_i\| = \|\sqrt{\tilde{\Sigma}} \begin{bmatrix}
H_i \\
\tilde{H}_i
\end{bmatrix}\|$ and the results of the previous section give that the minimizer of $f(\gamma_1,\gamma_2,...,\gamma_p)$ is a permutation of the elements in the vector $(\sigma_1,\sigma_2,...,\sigma_r,0,...,0)$.

We therefore have the following result:
\begin{corollary}
Let $f$ be quasi-concave (and lower semi-continuous) on $\mathbb{R}^p_{\geq 0}$ fulfilling $f(\tilde{\bfgamma}) \geq f(\bfgamma)$ when
$\tilde{\gamma}_i \geq \gamma_i$ for all $i$. Then an optimizer $\bfgamma^*$ of \eqref{eq:optproblrec}-\eqref{eq:optconstrec} is of the form  $\bfgamma^* = \Pi_{p \times r} \bfsigma$ where $\Pi_{p \times r}$ contains the first $r$ columns of a $p \times p$ permutation matrix.
\end{corollary}

\section{Linear Objectives - Weighted Nuclear Norms} \label{sec:wnn}

We now consider weighted nuclear norm regularization  $f(\bfgamma) = a^T \bfgamma$.
To ensure that the problem is well posed we assume that the elements of $a$ are non-negative. It is then clear that $f(\tilde{\bfgamma}) \geq f(\bfgamma)$ when $\tilde{\gamma}_i \geq \gamma_i$.
Since linearity implies concavity the results of Sections~\ref{sec:extremepts} and \ref{sec:generalmat} now show that the minimum of $f(M\bfsigma)$, over $M\in\mathcal{S}$ is attained in $M = \Pi$ for some permutation matrix.
To ensure that the bilinear formulation is equivalent to the original one we need to show that the optimum occurs when $\Pi = I$. 
Suppose that the elements in $a$ are sorted in ascending order $a_1 \leq a_2 \leq ... \leq a_p$.
It is easy to see that for $\Pi$ to give the smallest objective value it should sort the elements of $\gamma$ so that $\gamma_1 \geq \gamma_2 \geq ... \geq \gamma_p$, which means that $\Pi = I$ and $\bfgamma=\bfsigma$. We therefore conclude that minimizing \eqref{eq:generalbilinear} with a linear objective corresponds to regularization with a weighted nuclear norm with non-decreasing weights.

\section{Experiments}
In this section we start by describing implementation details of our method and then apply it to the problems of low matrix recovery and non-rigid structure recovery.
Solving the weighted nuclear norm regularized problem \eqref{eq:wnnbilinear} now amounts to minimizing
{\small
\begin{equation}
\sum_{i=1}^p a_i \frac{\|B_i\|^2+\|C_i\|^2}{2} + \|\A(BC^T) - b\|^2. 
\label{eq:fixweightedproblem}
\end{equation}}
Note that the terms in the objective \eqref{eq:fixweightedproblem} can be combined into a single norm term by vertically concatenating the vectors $B_i$ and $C_i$, weighted by $\sqrt{a_i/2}$, with $\A(BC^T) - b$. We define the resulting vector as $r_a := \A_a(BC^T) - b_a$, giving the objective  $\|r_a(BC^T)\|^2$, where the subscript reflects the dependence on the weights $a$.
Since the objective is smooth, standard methods such as Levenberg-Marquardt can be applied and Algorithm~\ref{alg:bilinear_param_wnn} shows an overview of the method used. Additional information about the algorithm is provided in the supplementary material.

The remainder of this section is organized as follows.
The particular form of the data fitting term in \eqref{eq:fixweightedproblem} when applied to structure from motion is described in Section~\ref{sec:pose_error}.
In Section~\ref{sec:exp_matrix_recovery} we compare the convergence of first and second-order methods, and motivated by the ADMM fast iterations but low accuracy, as opposed to the bilinear parameterization's high accuracy but slower iterations, we combine the two methods by initializing the bilinear parameterization with the ADMM's solution \cite{boyd-etal-2011,kumar-arxiv-2019} for a non-rigid structure structure recovery problem. Our work focus on the increased accuracy of our method compared to first-order methods, so the comparison of our results with  works such as \cite{xiao_2004,yan_2008,akhter_2008,paladini_2009} (without the desired regularization term) are not covered.
%Therefore VarPro can alternatively be applied as described in \cite{hong-fitzgibbon-cvpr-2015}.

\begin{algorithm}[t]
\algsetup{linenosize=\small}
  \scriptsize
\SetAlgoLined
\KwResult{Optimal B,C to \eqref{eq:fixweightedproblem}}
 $B = U\Sigma^{\frac{1}{2}}$, $C = V\Sigma^{\frac{1}{2}}$, where $X = U \Sigma V^T$ is the ADMM solution to \eqref{eq:fixweightedproblem} \;
 Choose initial $\alpha > 1$ and $\lambda$, and define $z = [\mathrm{vec}(B);\text{ }\mathrm{vec}(C^T)]$\;
 Compute \textit{error} $=\|r_a(BC^T)\|^2$\;
 \While{not converged}{
 Compute $r = r_a(BC^T)$, and the jacobian $J$ of \eqref{eq:fixweightedproblem} in terms of $z$\;
  Update $\tilde{z} = z - (J^TJ + \lambda I)^{-1} J^Tr$, where $\tilde{z} = [\mathrm{vec}(\tilde{B});\text{ }\mathrm{vec}(\tilde{C}^T)]$\;
  \eIf{error $>\|r_a(\tilde{B}\tilde{C}^T)\|^2$ }{
  Updata $z \leftarrow \tilde{z}$, 
   $\lambda \leftarrow \alpha^{-1} \lambda$, and 
   \textit{error} $\leftarrow \|\A_a(\tilde{B}\tilde{C}^T) - b_a\|^2$\;
   }{
   $\lambda \leftarrow \alpha \lambda$\;
  }
 }
 \caption{Bilinear parameterization of weighted nuclear norm}
 \label{alg:bilinear_param_wnn}
\end{algorithm}

\subsection{Pseudo Object Space Error (pOSE)  and Non-Rigid Structure from Motion}
\label{sec:pose_error}

To compare the performance of first and second-order methods, we choose as objective function the Pseudo Object Space Error (pOSE) \cite{hong-zach-cvpr-2018}, which consists of a combination of the object space error $\ell_{\mathrm{OSE}} := \sum_{(i,j)\in \Omega}\| \mathrm{P}_{i,1:2}\tilde{\bm{\mathrm{x}}}_j - (\bm{\mathrm{p}}_{i,3}^T\tilde{\bm{\mathrm{x}}}_j)\bm{\mathrm{m}}_{i,j} \|^2_2$
%\begin{equation}
%    \ell_{\mathrm{OSE}} := \sum_{(i,j)\in \Omega}\| \mathrm{P}_{i,1:2}\tilde{\bm{\mathrm{x}}}_j - (\bm{\mathrm{p}}_{i,3}^T\tilde{\bm{\mathrm{x}}}_j)\bm{\mathrm{m}}_{i,j} \|^2_2
%\end{equation}
and the affine projection error
%\begin{equation}
    $\ell_{\mathrm{Affine}} := \sum_{(i,j)\in \Omega}\| \mathrm{P}_{i,1:2}\tilde{\bm{\mathrm{x}}}_j - \bm{\mathrm{m}}_{i,j} \|^2_2$,
%\end{equation}
where $\mathrm{P}_{i,1:2}$ and $\bm{\mathrm{p}}_{i,3}$ are, respectively, the first two and the third rows of the camera matrix $\mathrm{P}_{i}$, with $i = 1,\hdots,F$; $\tilde{\bm{\mathrm{x}}}_j$ is a 3D point in homogeneous coordinates, with $j = 1,\hdots,P$; $\bm{\mathrm{m}}_{i,j}$ is the 2D observation of the $j$:th point on the $i$:th camera; and $\Omega$ represents the set of observable data. The Pseudo Object Space Error is then given by 
\begin{equation}
    \ell_{\mathrm{pOSE}} := (1-\eta)\ell_{\mathrm{OSE}} + \eta \ell_{\mathrm{Affine}}
\end{equation}
where $\eta \in [0,1]$ balances the weight between the two errors. One of the main properties of pOSE is its wide basin of convergence
while keeping a bilinear problem strucuture. 
The $\ell_{\mathrm{pOSE}}$, originally designed for rigid structure from motion, can be extended for the non-rigid case by replacing $\mathrm{P}_i\tilde{\bm{\mathrm{x}}}_j$ by a linear combination of $K$ shape basis, i.e.,  $\mathrm{\Pi}_i\hat{\mathrm{S}}_j$, where $\mathrm{\Pi}_i \in \mathbb{R}^{3 \times (3K+1)}$ and $\hat{\mathrm{S}}_j \in \mathbb{R}^{3K+1}$ are structured as
\begin{equation}
    \mathrm{\Pi}_i = \begin{bmatrix}
    c_{i,1}R_i \quad \hdots \quad c_{i,K}R_i \quad t_i
    \end{bmatrix} \quad \text{and} \quad 
    \hat{\mathrm{S}}_j = \begin{bmatrix}
    S_{1,j}^T \quad \hdots \quad S_{K,j}^T \quad 1
    \end{bmatrix}^T.
    \label{eq:nonrigid_fact}
\end{equation}
We denote by $\mathrm{\Pi}$ and $\hat{\mathrm{S}}$ the vertical and horizontal concatenations of $\mathrm{\Pi}_i$ and $\hat{\mathrm{S}}_j$, respectively. 
Note that for $K = 1$, we have $\mathrm{\Pi}_i\hat{\mathrm{S}}_j = \mathrm{P}_i\tilde{\bm{\mathrm{x}}}_j$, which corresponds to the rigid case, and consequentely $\mathrm{rank}(\mathrm{\Pi}\hat{\mathrm{S}}) = 4$. For $K>1$, by construction we have $\mathrm{rank}(\mathrm{\Pi}\hat{\mathrm{S}}) \leq 3K+1$.

\subsection{Low-Rank Matrix Recovery with pOSE errors}
\label{sec:exp_matrix_recovery}
In this section we compare the performance of first and second-order methods in terms of convergence and accuracy, starting from the same initial guess, for low-rank matrix recovery with pOSE. In this problem, we define $X = \mathrm{\Pi}\hat{\mathrm{S}}$ and aim at minimizing
\begin{equation}
\label{prob:low_rank_matrix_recovery}
\min_X a^T \bfsigma(X) + \ell_{\mathrm{pOSE}}(X),
\end{equation}We apply our method and solve the optimization problem defined in 
\eqref{eq:fixweightedproblem} by using the bilinear factorization $X = BC^T$, with $B \in \mathbb{R}^{3F \times r}$, and $C \in \mathbb{R}^{P \times r}$, with $r \geq 3K+1$. We test the performance of our method in 4 datasets: Door \cite{olsson-engqvist-scia-2011}, Back \cite{Russell2011}, Heart \cite{stoyanov2005}, Paper \cite{Varol2009}. The first one consists of image measurements of a rigid structure with missing data, while the remaining three datasets track points in deformable structures. 

For the Door dataset, we apply two different selections of weights on the singular values of $X$, corresponding to the nuclear norm, i.e., $a_i = \mu_{NN}$, and truncated nuclear norm, i.e., $a_i = 0$, $i = 1,\hdots,4$ and $a_i = \mu_{TNN}$, $i > 4$. We select $\mu_{NN} = 1.5\times 10^{-3}$, and $\mu_{TNN} = 1$.
For the Back, Heart and Paper datasets, we apply the nuclear norm and a weighted nuclear norm, in which the first four singular values of $X$ are not penalized and the remaining ones are increasingly penalized, i.e., $a_i = 0$, $i = 1,\hdots,4$ and $a_i = (i-4)\mu_{WNN}$, $i > 4$. 
We select $\mu_{NN} = 7.5\times 10^{-4}$, $\mu_{WNN} = 2.25\times 10^{-3}$. The values of the weights $a_i$ are chosen such that there is a $3K + 1$ rank solution to \eqref{prob:low_rank_matrix_recovery}, with $K = 1$ and $K=2$ for the rigid and non-rigid datasets, respectively.  

\begin{figure}
\centering
\begin{subfigure}{\textwidth}
 \centering
\includegraphics[width=0.47\linewidth]{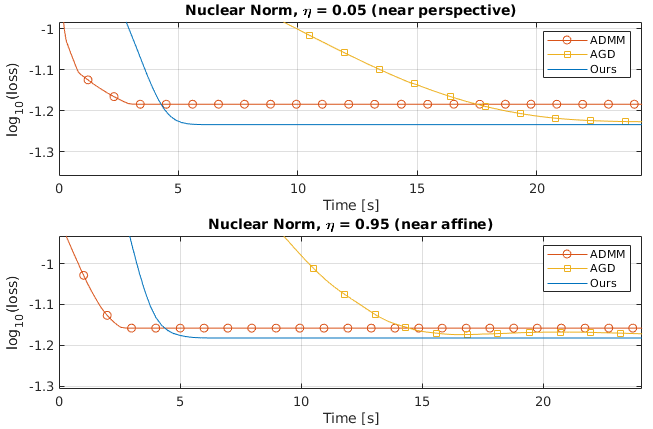}
\hfill
 \includegraphics[width=0.47\linewidth]{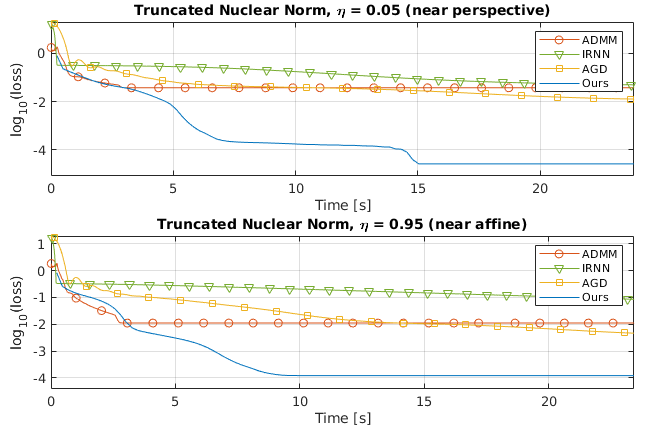}
\caption{Door Dataset.}
\centering
\end{subfigure}
\begin{subfigure}{\textwidth}
 \centering
\includegraphics[width=0.47\linewidth]{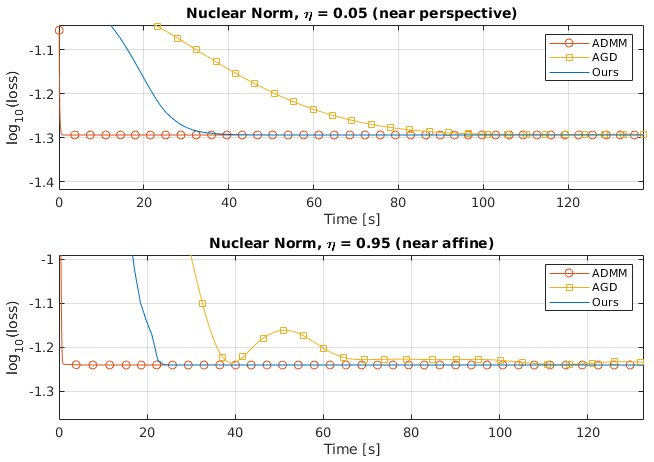}
\hfill
 \includegraphics[width=0.47\linewidth]{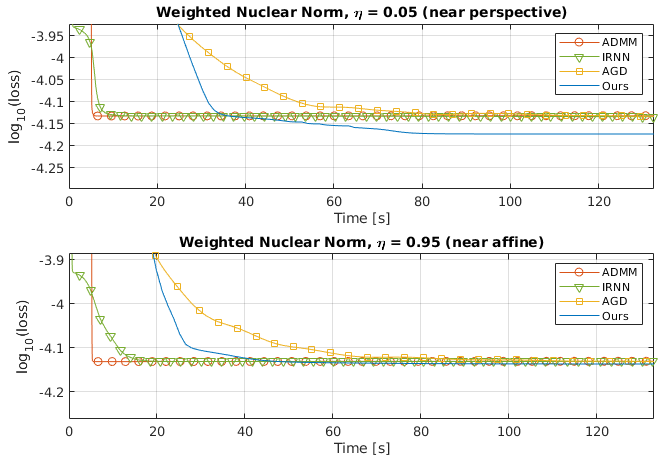}
\caption{Heart Dataset.}
\centering
\end{subfigure}
\begin{subfigure}{\textwidth}
 \centering
\includegraphics[width=0.47\linewidth]{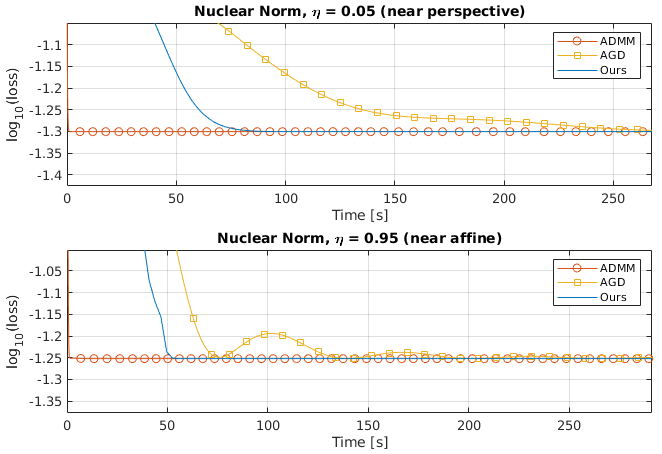}
\hfill
 \includegraphics[width=0.47\linewidth]{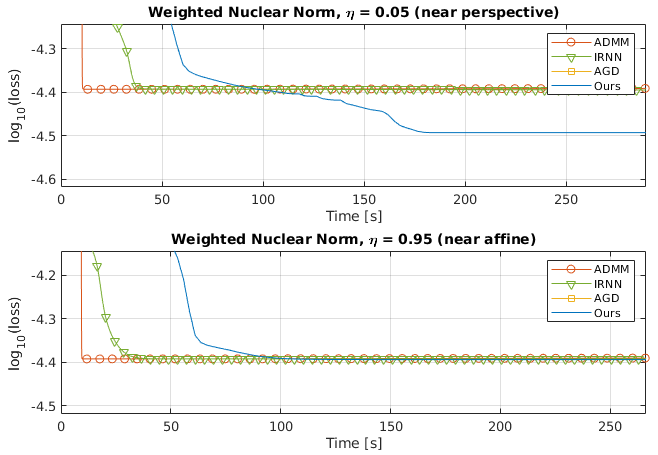}
\caption{Back Dataset.}
\centering
\end{subfigure}
\begin{subfigure}{\textwidth}
 \centering
\includegraphics[width=0.47\linewidth]{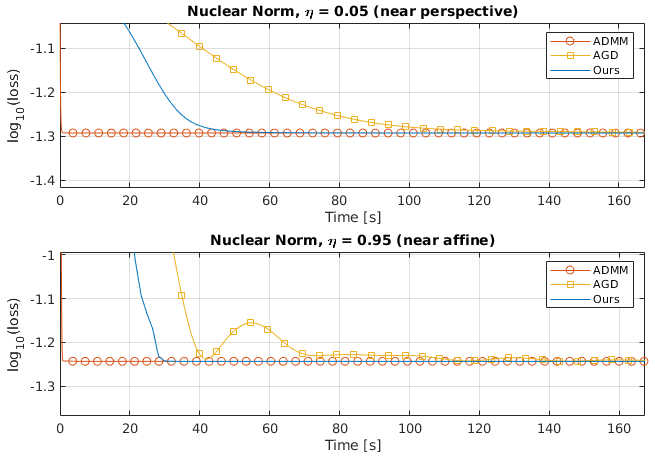}
\hfill
 \includegraphics[width=0.47\linewidth]{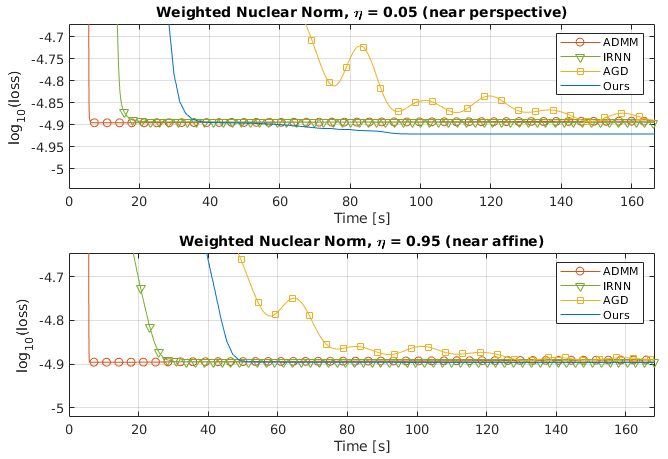}
\caption{Paper Dataset.}
\centering
\end{subfigure}
\caption{Convergence of the four methods for low-rank matrix recovery on the Door, Heart, Back and Paper datasets. }
\label{fig:exps_matrix_recovery_loss}
\end{figure}

We compare the bilinear parameterization with three first-order methods commonly used for low-rank matrix recovery: Alternating Direction Method of Multipliers (ADMM) \cite{boyd-etal-2011}, Iteratively Reweighted Nuclear Norm (IRNN) \cite{canyi2015}, and Accelerated Gradient Descend (AGD) \cite{li2019}. We also test the methods for two different cases of the $\ell_{\mathrm{pOSE}}$ error, with $\eta = 0.05$ and $\eta = 0.95$, which correspond to the near-perspective and near-affine camera models, respectively. To improve numerical stability of the algorithms, as pre-processing step we normalize the image measurements matrix $M$ by its norm. The methods are initialized with the closed-form solution of the regularization-free problem, i.e., $X = \A^\dagger(b)$. The comparison of the four algorithms in terms of total log-loss over time is shown in Figure~\ref{fig:exps_matrix_recovery_loss}. The log-loss is used for better visualization purposes. The plots for the IRNN for the nuclear norm are omitted since it demonstrated slow convergence compared to the remaining three methods. A qualitative evaluation of the results on one of the images of the Door dataset for the truncated nuclear norm and near perspective camera model is shown in Figure~\ref{fig:door_eval}.
The qualitative results for the remaining datasets are provided in the supplementary material.%The same evaluation for the other cases is not shown since the difference between the methods is not noticeable. 

In general, we can observe that first-order methods demonstrate faster initial convergence, mostly due to faster iterations. However when near minima, the convergence rate drops significantly and the methods tend to stall. Contrarily, bilinear parameterization compensates its slower iterations by demonstrating higher accuracy and and reaching solutions with lower energy. This is specially visible for the near perspective camera model, which reinforces the advantages of using a second-order method on image data under perspective projection. To compensate for the slower convergence, we propose the initialization of the bilinear parameterization with the solution obtained with ADMM. In this way, the bilinear parameterization starts near the minimum and performs local refinement to further improve accuracy.  

%The experiments with the nuclear norm, the convex formulation, show the difference in terms of speed of convergence of the methods, with the first order methods showing slow convergence compared to the bilinear factorization, with the exception of ADMM for the non-rigid datasets. In these datasets, ADMM demonstrates extremely fast convergence independently of the camera model. However, when we move to the non-convex formulation with the truncated and weighted nuclear norms, all first order methods fail at achieving minima as good as the bilinear factorization. The gap between the losses obtained by the bilinear factorization and the first order methods is larger for the near perspective camera model, and it is attenuated as we approach the affine model.

\begin{figure}
\centering
\begin{subfigure}{\textwidth}
 \centering
\includegraphics[height=3.5cm]{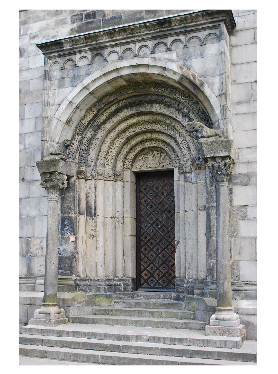}
\includegraphics[width=0.19\linewidth]{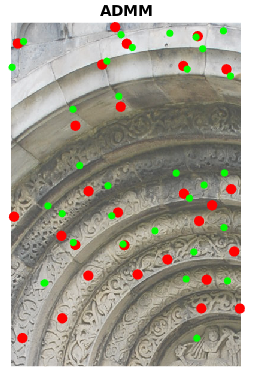}
\includegraphics[width=0.1925\linewidth]{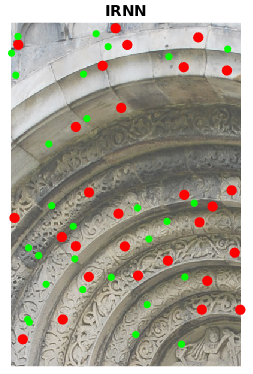}
\includegraphics[width=0.19\linewidth]{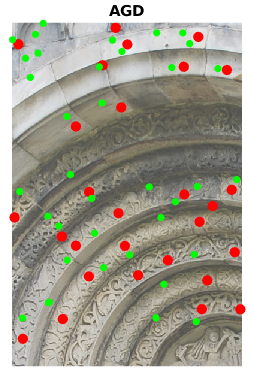}
\includegraphics[width=0.19\linewidth]{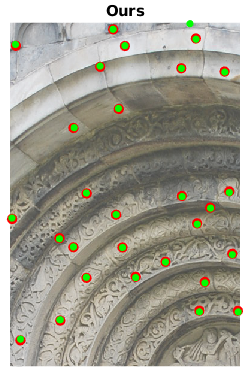}
\centering
\end{subfigure}
\caption{Evaluation of the four methods for low-rank matrix recovery on one of the images of the Door dataset. The red circles show the target image measurements and the green circles the estimate image points. }
\label{fig:door_eval}
\end{figure}

\begin{figure}
\centering
\begin{subfigure}{0.15\textwidth}
\end{subfigure}
\begin{subfigure}{0.03\textwidth}
\rotatebox[origin=t]{90}{Articulated}
\end{subfigure}
\begin{subfigure}{0.78\textwidth}
\hspace{.25cm}
\includegraphics[height=2.275cm]{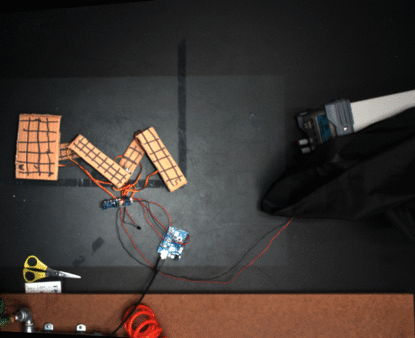}\hspace{1.5cm}
\includegraphics[height=2.25cm]{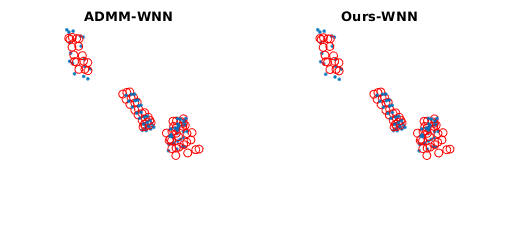}
\end{subfigure}

\begin{subfigure}{0.15\textwidth}
\end{subfigure}
\begin{subfigure}{0.03\textwidth}
\rotatebox[origin=t]{90}{Balloon}
\end{subfigure}
\begin{subfigure}{0.78\textwidth}
\hspace{.235cm}
\includegraphics[height=2.275cm]{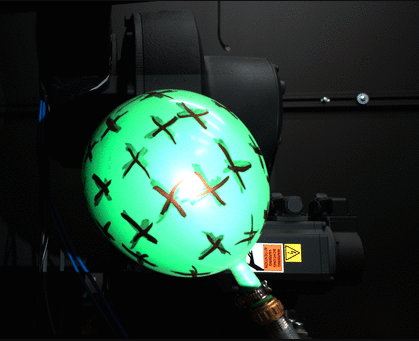}
\hspace{1.5cm}\includegraphics[height=2.25cm]{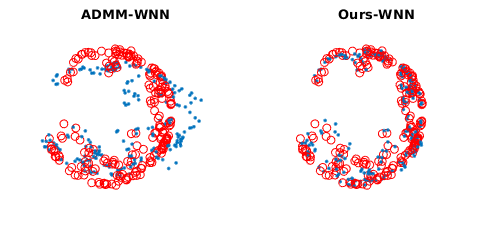}
\end{subfigure}

\begin{subfigure}{0.15\textwidth}
\end{subfigure}
\begin{subfigure}{0.03\textwidth}
\rotatebox[origin=t]{90}{Paper}
\end{subfigure}
\begin{subfigure}{0.78\textwidth}
\hspace{.325cm}\includegraphics[height=2.25cm]{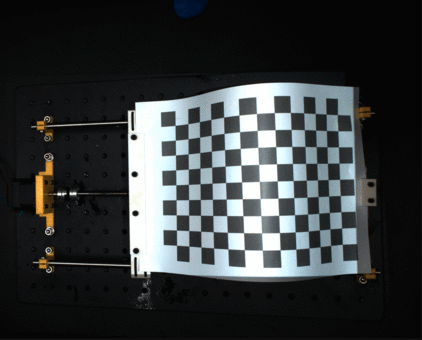}
\hspace{1.5cm}\includegraphics[height=2.25cm]{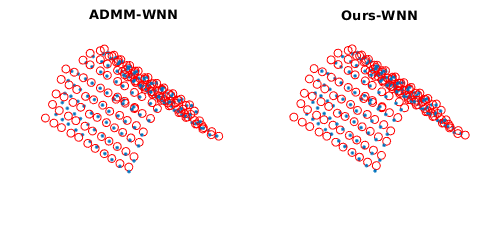}
\end{subfigure}

\begin{subfigure}{0.15\textwidth}
\end{subfigure}
\begin{subfigure}{0.03\textwidth}
\rotatebox[origin=t]{90}{Stretch}
\end{subfigure}
\begin{subfigure}{0.78\textwidth}
\hspace{.32cm}\includegraphics[height=2.25cm]{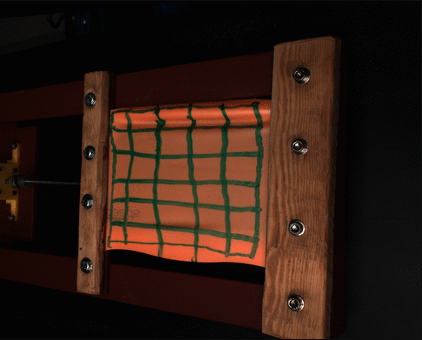}
\hspace{1.5cm}\includegraphics[height=2.25cm]{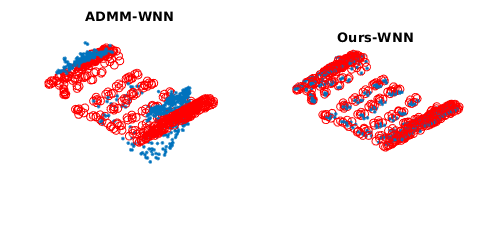}
\end{subfigure}

\begin{subfigure}{0.15\textwidth}
\end{subfigure}
\begin{subfigure}{0.03\textwidth}
\rotatebox[origin=t]{90}{Tearing}
\end{subfigure}
\begin{subfigure}{0.78\textwidth}
\hspace{.325cm}\includegraphics[height=2.25cm]{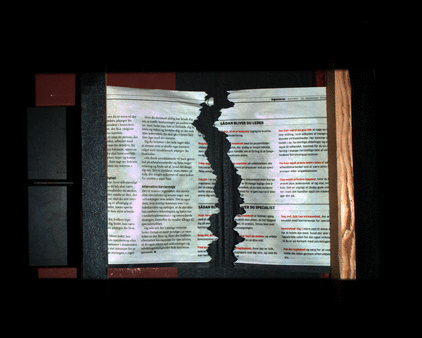}
\hspace{1.5cm}\includegraphics[height=2.25cm]{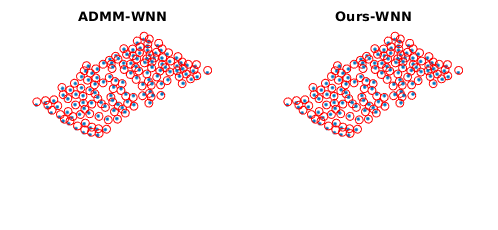}
\end{subfigure}
\caption{(Left) Example of the non-rigid objects in the 5 datasets of the NRSfM Challenge. (Right) Estimation (blue) and ground-truth (red) of the non-rigid 3D structure for the two methods with weighted nuclear norm regularization.}
\label{fig:nrsfm_results}
\end{figure}

\iffalse
\begin{figure}
\centering
\includegraphics[height=3.5cm]{images/hybrid.png}
\caption{
}
\end{figure}
\fi

\subsection{Non-Rigid Structure Recovery}
Consider now that the camera rotations in \eqref{eq:nonrigid_fact} are known (or previously estimated). In this case we have $ \mathrm{\Pi}\hat{\mathrm{S}} = RX+t\mathbbm{1}^T$, with $R = \mathrm{blkdiag}(R_1,\hdots,R_F)$ and $t = [t_1^T,\hdots,t_F^T]^T$, where $X$, the non-rigid structure, and $t$ are the unknowns. It is directly observed that $\mathrm{rank}(\mathrm{\Pi}\hat{\mathrm{S}})\leq \mathrm{rank}(RX) + \mathrm{rank}(t\mathbbm{1}^T)$, with the later being equal to 1 by construction and independent on $K$. As consequence, it follows that $\mathrm{rank}(RX) = \mathrm{rank}(X) \leq 3K$, and the rank regularization can be applied on $X$. A similar problem was studied in \cite{dai-etal-ijcv-2014} but for orthogonal camera models, where the authors propose the rank regularization to be applied on a reshaped version of $X$, a $F \times 3P$ matrix structured as
{\small
\begin{equation}
    X^\# = \begin{bmatrix} X_{1,1} & \hdots & X_{1,P} & Y_{1,1} & \hdots & Y_{1,P} & Z_{1,1} & \hdots & Z_{1,P}  \\
    \vdots & & \vdots & \vdots & & \vdots & \vdots & & \vdots \\
    X_{F,1} & \hdots & X_{F,P} & Y_{F,1} & \hdots & Y_{F,P} & Z_{F,1} & \hdots & Z_{F,P}
    \end{bmatrix} = g^{-1}(X),
\end{equation}}\normalsize
where $X_{i,j}$, $Y_{i,j}$ and $Z_{i,j}$ are respectively the x, y and z-coordinates of the $j$:th 3D point at frame $i$. The function $g$ performs the permutation on the elements of $X^\#$ to obtain $X$. With this reshaping we have that $\mathrm{rank}(X^\#) \leq K$, meaning that we can factorize it as $X^\# = BC^T$ with $B \in \mathbb{R}^{F \times K}$ and $C \in \mathbb{R}^{3P \times K}$. The optimization problem then becomes 
{\small
\begin{equation}
\min_{B,C,t} \sum_{i=1}^K a_i \frac{\|B_i\|^2+\|C_i\|^2}{2} + \ell_{\mathrm{pOSE}}(Rg(BC^T)+t\mathbbm{1}^T). 
\label{eq:nonrigid_structure_opt}
\end{equation}}Solving this optimization problem requires small adjustments to be done to the proposed Algorithm~\ref{alg:bilinear_param_wnn}, which can be consulted in the supplementary material.
We apply our methods to the 5 datasets (Articulated, Balloon, Paper, Stretch, Tearing) from the NRSfM Challenge \cite{jensen2018}. Each of these datasets include tracks of image points for orthogonal and perspective camera models for six different camera paths (Circle, Flyby, Line, Semi-circle, Tricky, Zigzag), as well as the ground-truth 3D structure for one of the frames. We use the 2D observation for the orthogonal camera model to compute the rotation matrix $R$, as done in \cite{dai-etal-ijcv-2014}, and the ground-truth 3D structure to estimate the intrinsic camera matrix, which is assumed to be fixed during each sequence. The intrinsic camera matrix is used to obtain the calibrated 2D observation of the perspective camera model data. For the nuclear norm (NN), we set $a_i = 1\times 10^{-3}, i = 1,\hdots,K$. For the weighted nuclear norm (WNN), the weights $a$ are selected similarly to \cite{kumar-arxiv-2019}
{\small
\begin{equation}
    a_i = \frac{\xi}{\sigma_i(g^{-1}(X_0)) + \gamma}, \quad i = 1,\hdots,K
\end{equation}}
where $\xi = 5\times 10^{-3}$, $\gamma$ is a small number for numerical stability, and $X_0$ is the closed-form solution of the objective $\min_{X}\ell_{\mathrm{pOSE}}(RX). $

For these datasets we choose $K = 2$ and set the $\eta = 0.05$. As baseline we use the best performing first-order method according to the experiments Section~\ref{sec:exp_matrix_recovery}, ADMM, and apply the method described in Algorithm~\ref{alg:bilinear_param_wnn} for local refinement starting from the ADMM's solution. We also try our method for the orthogonal camera model (by setting $\eta = 1$), and compare it with BMM \cite{dai-etal-ijcv-2014} and R-BMM \cite{kumar-arxiv-2019}, which correspond to ADMM implementations for nuclear norm and weighted nuclear norm, respectively. These methods perform a best rank $K$ approximation to the obtained ADMM solution if $\mathrm{rank}(X^\#) > K$ after convergence. We let the ADMM-based methods run until convergence or stalling is achieved for fair comparison. The average log-losses, before and after refinement, obtained on each dataset over the 6 camera paths are shown in Table~\ref{tab:nrsfm_data_results_loss}. The average reconstruction errors, in millimeters, on each dataset over the 6 paths relatively to the provided ground-truth structure are shown in Table~\ref{tab:nrsfm_data_results}. In Figure~\ref{fig:nrsfm_results} we also show some qualitative results of the obtained 3D reconstruction of each of the objects in the 5 datasets. More qualitative results are provided in the supplementary material.

The results show that our method is able to achieve lower energies for all datasets comparatively with the ADMM baselines. As expected from the experiments in Section~\ref{sec:exp_matrix_recovery}, the difference is more substantial for the perspective model. Furthermore, even though we are not explicitly minimizing the reconstruction error expressed in Table~\ref{tab:nrsfm_data_results}, we are able to consistently obtain the lowest reconstruction error for all datasets, sometimes with great improvements compared to the ADMM (see Balloon and Stretch in Figure~\ref{fig:nrsfm_results}). The same does not apply for the orthogonal data, where achieving lower energies did not lead to lower reconstruction errors.

\small{
\begin{table}[t]
\caption{Average log-loss on each of the perspective datasets over the 6 camera paths.}
\label{tab:nrsfm_data_results_loss}
\begin{tabular}{>{\centering}p{0.13\textwidth}>{\centering}p{0.27\textwidth}>{\centering}p{0.12\textwidth}>{\centering}p{0.12\textwidth}>{\centering}p{0.12\textwidth}>{\centering}p{0.12\textwidth}>{\centering\arraybackslash}p{0.12\textwidth}}
  &\multicolumn{1}{c|}{Method \textbackslash Dataset} & Articulated & Balloon & Paper & Stretch & Tearing \\ \cline{2-7} 
\multirow{4}{*}{Orthogonal} 
%& \multicolumn{1}{c|}{Multibody \cite{kumar2017}} & 10.15 & 10.64 & 15.78 & 9.96 & 14.17 \\
 & \multicolumn{1}{c|}{BMM \cite{dai-etal-ijcv-2014}} & -1.645 & -2.267 & -1.712 & -2.282 & -1.453 \\
   &                     \multicolumn{1}{c|}{Ours-NN} & \textbf{-1.800} & \textbf{-2.352} & \textbf{-2.188} & \textbf{-2.509} & \textbf{-1.634} \\\cdashline{2-7}
  & \multicolumn{1}{c|}{R-BMM \cite{kumar-arxiv-2019}} & -1.648 & -1.979 & -1.855 & -1.997 & -1.522 \\
  %& \multicolumn{1}{c|}{Ours (MD)} & TBA & TBA & TBA & TBA & TBA\\
   & \multicolumn{1}{c|}{Ours-WNN} & \textbf{-1.648} & \textbf{-1.979} & \textbf{-1.855} & \textbf{-1.997} & \textbf{-1.522} \\\cline{2-7}
   \multirow{4}{*}{Perspective} 
&\multicolumn{1}{c|}{ADMM-NN} & -2.221 & -2.529 & -2.338 & -2.395 & -1.471 \\
  &\multicolumn{1}{c|}{Ours-NN} & \textbf{-2.415} & \textbf{-2.657} & \textbf{-2.560} & \textbf{-2.622} & \textbf{-2.053} \\\cdashline{2-7}
 %& \multicolumn{1}{c|}{ADMM-WNN-MD} & -2.487 & -3.285 & -3.287 & -2.728 & -2.149 \\
  %& \multicolumn{1}{c|}{Ours (MD)} & TBA & TBA & TBA & TBA & TBA\\
  %&\multicolumn{1}{c|}{Ours-WNN-MD} & \textbf{-2.488} & \textbf{-3.309} & \textbf{-3.308} & \textbf{-2.915} &  \textbf{-2.166}\\\cdashline{2-7}
 & \multicolumn{1}{c|}{ADMM-WNN} & -2.455 & -2.617 & -2.195 & -2.651 & -1.688 \\
  %& \multicolumn{1}{c|}{Ours (MD)} & TBA & TBA & TBA & TBA & TBA\\
  &\multicolumn{1}{c|}{Ours-WNN} & \textbf{-2.486} & \textbf{-2.931} & \textbf{-2.777} & \textbf{-2.857} &  \textbf{-2.103}
\end{tabular}
\end{table}}
\tiny{
\begin{table}[t]
\caption{Average reconstruction errors, in millimeters, on each dataset over the 6 camera paths relatively to the provided ground-truth structure.}
\label{tab:nrsfm_data_results}
\begin{tabular}{>{\centering}p{0.13\textwidth}>{\centering}p{0.27\textwidth}>{\centering}p{0.12\textwidth}>{\centering}p{0.12\textwidth}>{\centering}p{0.12\textwidth}>{\centering}p{0.12\textwidth}>{\centering\arraybackslash}p{0.12\textwidth}}
 & \multicolumn{1}{c|}{Method \textbackslash Dataset} & Articulated & Balloon & Paper & Stretch & Tearing \\ \cline{2-7} 
\multirow{4}{*}{Orthogonal} 
%& \multicolumn{1}{c|}{Multibody \cite{kumar2017}} & 10.15 & 10.64 & 15.78 & 9.96 & 14.17 \\
 & \multicolumn{1}{c|}{BMM \cite{dai-etal-ijcv-2014}} & 18.49 & 10.39 & \textbf{8.94} & 10.02 & \textbf{14.23}\\
   & \multicolumn{1}{c|}{Ours-NN} & 18.31 & 8.53 & 10.94 & 10.67 & 17.03 \\
  & \multicolumn{1}{c|}{R-BMM \cite{kumar-arxiv-2019}} & 16.00 & \textbf{7.84} & 10.69 & \textbf{7.53} &  16.34 \\
  %& \multicolumn{1}{c|}{Ours (MD)} & TBA & TBA & TBA & TBA & TBA\\
   & \multicolumn{1}{c|}{Ours-WNN} & \textbf{15.03} & 8.05 & 10.45 & 9.01 & 16.20 \\\cline{2-7}
\multirow{4}{*}{Perspective} 
& \multicolumn{1}{c|}{ADMM-NN} & 16.70 & 8.05 & 7.96 & 6.04 & 9.40 \\
  %& \multicolumn{1}{c|}{Ours (MD)} & TBA & TBA & TBA & TBA & TBA\\
 & \multicolumn{1}{c|}{Ours-NN} & \textbf{16.13} & 6.48 & 6.80 & 6.00 & 9.31 \\
 %& \multicolumn{1}{c|}{ADMM-WNN-MD} & 18.18 & 10.52 & 10.92 & 8.70 & \textbf{9.00} \\
  %& \multicolumn{1}{c|}{Ours (MD)} & TBA & TBA & TBA & TBA & TBA\\
 %& \multicolumn{1}{c|}{Ours-WNN-MD} & 16.90 & 10.42 & 9.55 & 6.15 & 9.76 \\
 & \multicolumn{1}{c|}{ADMM-WNN} & 18.33 & 8.95 & 10.14 & 8.06 & 9.28 \\
  %& \multicolumn{1}{c|}{Ours (MD)} & TBA & TBA & TBA & TBA & TBA\\
 & \multicolumn{1}{c|}{Ours-WNN} & 16.53 & \textbf{6.27} & \textbf{5.68} & \textbf{5.93} & \textbf{8.42}
\end{tabular}
\end{table}}
\normalsize

\section{Conclusions}
In this paper we showed that it is possible to optimize a general class of singular value penalties using a bilinear parameterization of the matrix. We showed that with this parameterization weighted nuclear norm penalties turn in to smooth objectives that can be accurately solved with 2nd order methods. Our proposed approach starts by using ADMM which rapidly decreases the objective the first couple of iterations and switches to Levenberg-Marquardt when ADMM iterations make little progress. This results in a much more accurate solution and we showed that we were able to extend the recently proposed pOSE \cite{hong-zach-cvpr-2018} to handle non-rigid reconstruction problems.

While 2nd order methods offer increased accuracy, our approach is expensive since iterations require the inversion of a large matrix. Exploring feasible alternatives such as preconditioning and conjugate gradient approaches is an interesting future direction.

Something that we have not discussed is adding constraints on the factors, which is possible since these are present in the optimization. This is very relevant for structure from motion problems and will likely be an fruitful direction to explore.

\iffalse
\begin{figure}
\centering
\begin{subfigure}{0.5\textwidth}
\includegraphics[width=\textwidth]{images/door_detail.png}
\centering
  \label{fig:door_example}
\end{subfigure}
\begin{subfigure}{0.8\textwidth}
 \includegraphics[width=\textwidth]{images/heart_example.png}
\centering
  \label{fig:heart_example}
\end{subfigure}
\begin{subfigure}{0.8\textwidth}
 \includegraphics[width=\textwidth]{images/back_example.png}
\centering
  \label{fig:back_example}
\end{subfigure}
\begin{subfigure}{0.8\textwidth}
 \includegraphics[width=\textwidth]{images/paper_example.png}
\centering
  \label{fig:paper_example}
\end{subfigure}
\caption{plots of....}
\label{fig:fig}
\end{figure}
\fi

\iffalse
\begin{center}
\begin{tabular}{ |c |c| c| c |c| c| c| }
\hline
 Method & PTA & CSF2 & PND & BMM & Kumar & Ours \\ 
 \hline
 Drink & 0.0287 & 0.0227 & 0.0037 & 0.0266 & 0.0119 & 0.0366 \\  
\hline
 Pickup & 0.1939 & 0.1791 & 0.0372 & 0.1731 & 0.0198 & 0.0717 \\ 
 \hline
 Yoga & 0.1243 & 0.1179 & 0.0140 & 0.1150 & 0.0129 & 0.0414 \\ 
 \hline
 Stretch & 0.1035 & 0.1136 & 0.0156 & 0.1034 & 0.0144 & 0.0795 \\ 
 \hline
 Dance & 0.2426 & 0.1877 & 0.1454 & 0.1864 & 0.1060 & ? \\ 
 \hline
 Walking & 0.3761 & 0.1938 & 0.0465 &  0.1298 & 0.0882 & ? \\ 
 \hline
 Face & 0.0489 & 0.0319 & 0.0165 & 0.0303 & 0.0179 & ? \\ 
 \hline
 Shark & 0.2933 & 0.1117 & 0.0135 & 0.2311 & 0.0551 & ? \\
 \hline
\end{tabular}
\end{center}
\fi

\small
%\bibliographystyle{plain}

\begin{comment}
\documentclass[runningheads]{llncs}
\usepackage{graphicx}
\usepackage{comment}
\usepackage{amsmath,amssymb,bm} % define this before the line numbering.
\usepackage{color}

\usepackage[utf8]{inputenc}
%\usepackage{amsthm}
\usepackage{natbib}
\usepackage{graphicx}
\usepackage{epsfig}
\usepackage{caption}
\usepackage{subcaption}
\captionsetup{compatibility=false}
\usepackage{multirow}
\usepackage{array}
\usepackage{arydshln}
\usepackage[ruled,vlined]{algorithm2e}
\usepackage{algorithmic,float}
\usepackage{rotating}

% INITIAL SUBMISSION - The following two lines are NOT commented
% CAMERA READY - Comment OUT the following two lines
%\usepackage{ruler}
%\usepackage[width=122mm,left=12mm,paperwidth=146mm,height=193mm,top=12mm,paperheight=217mm]{geometry}

\end{comment}

%\begin{document}

\renewcommand\thesection{\Alph{section}}
\renewcommand\thesubsection{\thesection.\Roman{subsection}}
\numberwithin{equation}{section}
\counterwithin{table}{section}
\counterwithin{figure}{section}

\title{Supplementary Material for Accurate Optimization of Weighted Nuclear Norm for Non-Rigid Structure from Motion} % Replace with your title

\begin{comment}
\titlerunning{ECCV-20 submission ID \ECCVSubNumber} 
\authorrunning{ECCV-20 submission ID \ECCVSubNumber} 
\author{Anonymous ECCV submission}
\institute{Paper ID \ECCVSubNumber}
\end{comment}
%******************

%\begin{comment}
\titlerunning{Accurate Optimization of Weighted Nuclear Norm for NRSfM}
\author{José Pedro Iglesias\inst{1} \and
Carl Olsson\inst{1,2} \and
Marcus Valtonen Örnhag \inst{2}}
\authorrunning{J.P. Iglesias et al.}
\institute{Chalmers University of Technology, Sweden \and
Lund University, Sweden }
%\end{comment}
%******************
\maketitle

%===========================================================
\section{More details about Algorithm 1 and its implementation}

In the main paper we propose the minimization of the objective
\begin{equation}
\label{eq:sup_obj}
    \min_X a^T \bfsigma(X) + \|\A X - b\|^2
\end{equation}
through a bilinear parameterization $X = BC^T$ and using second-order optimization methods such as   Levenberg-Marquardt. In Section 5 we provide an overview of the algorithm used, and in this section of the supplementary material we provide more details that were omitted from the main text, in particular how to formulate the problems regarding Low-rank Matrix Recovery (Section 5.2) and Non-rigid Structure Recovery (Section 5.3) using the pOSE error introduced in Section 5.1.

We start by showing how the pOSE term in \eqref{eq:sup_obj} can be written as linear mapping of the elements of $X$, resulting in the equivalent objective 
\begin{equation}
\label{eq:sup_obj2}
    \min_X a^T \bfsigma(X) + \| A_X\vec{X} - b_X\|^2.
\end{equation}
The terms $\ell_{\mathrm{Affine}}$ and $\ell_{\mathrm{OSE}}$ of the pOSE can be written as 
\begin{equation}
\label{eq:l_affine_vec}
    \ell_{\mathrm{Affine}} = \|\Gamma_{1:2}X - M\|_F^2 = \|(I \otimes \Gamma_{1:2})\vec{X} - \mathbf{m}\|^2
\end{equation}
and
\begin{equation}
\label{eq:l_ose_vec}
    \ell_{\mathrm{OSE}} = \|\Gamma_{1:2}X - \Gamma_3X \odot M\|^2_F = \|\left((I \otimes \Gamma_{1:2}) - \diag(\mathbf{m})(I \otimes \Gamma_3)\right)\vec{X} \|^2,
\end{equation}
where the matrices $\Gamma_{1:2},\Gamma_3 \in \mathbbm{R}^{2F \times 3F}$ select the desired rows of $X$ and $M \in \mathbbm{R}^{2F \times P}$ gathers all the 2D observations $\bm{\mathrm{m}}_{i,j}$ with $i =1,\hdots,F$ and $j = 1,\hdots,P$. We define $\bm{\mathrm{m}} = \vec{M}$. The rows $2i-1$ and $2i$ of $\Gamma_{1:2}X$ are equal to the rows $3i-2$ and $3i-1$ of $X$, respectively. The rows $2i-1$ and $2i$ of $\Gamma_3X$ are both equal to the row $3i$ of $X$. To obtain \eqref{eq:l_affine_vec} and \eqref{eq:l_ose_vec} we use $\vec{AXB} = (B^T \otimes A)\vec{X}$, where $\otimes$ denotes the Kronocker product.

This allows us to write $A_X$ and $b_X$ in \eqref{eq:sup_obj2} as
\begin{equation}
    A_X = \begin{bmatrix} \sqrt{\eta}\left(I \otimes \Gamma_{1:2}\right)\\
    \sqrt{1-\eta}\left((I \otimes \Gamma_{1:2}) - \diag(\mathbf{m})(I \otimes \Gamma_3)\right)
    \end{bmatrix},\quad
    b_X = \begin{bmatrix}  \sqrt{\eta}\mathbf{m}\\
    \mathbf{0}
    \end{bmatrix}.
\end{equation}
We use this as starting point to formulate the problems of Low-rank Matrix Recovery and Non-rigid Structure Recovery, which differ on the way $X$ in \eqref{eq:sup_obj2} is parameterized. 

\subsection{Low-rank Matrix Recovery with pOSE errors}
As seen in Section 5.2, we parameterize $X = BC^T$, which results in the objective 
\begin{equation}
\sum_{i=1}^p a_i \frac{\|B_i\|^2+\|C_i\|^2}{2} + \|A_X\vec{BC^T} - b_X\|^2. 
\label{eq:sup_fixweightedproblem}
\end{equation}
The pOSE term in \eqref{eq:sup_fixweightedproblem} is no longer linear in $B$ and $C$, and in order to apply Levenberg-Marquardt method we linearize it in the neighbourhood of $B_0$ and $C_0$ as
\begin{equation}
     A_X\vec{BC^T} - b_X \approx \left(A_X\vec{B_0C_0^T} - b_X \right) + A_B\vec{\delta B} + A_{C^T}\vec{\delta C^T}
\end{equation}
where we define
\begin{equation}
    r_{\mathrm{pOSE}} = A_X\vec{B_0C_0^T} - b_X
\end{equation}
with
\begin{equation}
\label{eq:ab_ac_matrices}
    A_B = A_X(C_0 \otimes I), \quad A_{C^T} = A_X(I \otimes B_0).
\end{equation}
 The terms corresponding to the weighted nuclear norm can also be written in a similar fashion since we have
\begin{equation}
    \sum_i \frac{a_i}{2}\|B_i\|^2 = \|B\diag(\sqrt{a/2})\|_F^2 = \|(\diag(\sqrt{a/2}) \otimes I )\vec{B}\|^2,
\end{equation}
\begin{equation}
    \sum_i \frac{a_i}{2}\|C_i\|^2 = \|\diag(\sqrt{a/2})C^T\|_F^2 = \|(I \otimes \diag(\sqrt{a/2}))\vec{C^T}\|^2.
\end{equation}
Again, by considering the deviations from the current estimations, $B = B_0 + \delta B$ and $C = C_0 + \delta C$, we end up with 
\begin{equation}
\label{eq:sup_regB}
    A_{\mathrm{regB}} = \diag(\sqrt{a/2}) \otimes I, \quad r_{\mathrm{regB}} = (\diag(\sqrt{a/2}) \otimes I )\vec{B_0},
\end{equation}
\begin{equation}
\label{eq:sup_regC}
    A_{\mathrm{regC}} = I \otimes \diag(\sqrt{a/2}), \quad r_{\mathrm{regC}} = (I \otimes \diag(\sqrt{a/2}))\vec{C_0^T}.
\end{equation}
As so, we can compute the residuals and jacobian in Algorithm 1 for the Low-rank Matrix Recovery problem as 
\begin{equation}
    J = \begin{bmatrix} A_B & A_{C^T}\\
    A_{\mathrm{regB}} & \mathbf{0}\\
    \mathbf{0} & A_{\mathrm{regC}}
    \end{bmatrix}, \quad r_a = \A_a(B_0C_0^T) + b_a = \begin{bmatrix}
    r_{\mathrm{pOSE}}\\r_{\mathrm{regB}}\\r_{\mathrm{regC}}
    \end{bmatrix}.
\end{equation}

\subsection{Non-Rigid Structure Recovery}
When considering the Non-rigid Structure Recovery problem, we use the parameterization $X = Rg(BC^T)+t\mathbbm{1}^T$. This also results in a non-linear pOSE term in terms of $B$ and $C$, and its linearization around $B_0$, $C_0$ and $t_0$ are obtained as 
\begin{multline}
     A_X\vec{Rg(BC^T)+t\mathbbm{1}^T} - b_X \approx \\
     \approx \left(A_X\vec{Rg(B_0C_0^T)+t_0\mathbbm{1}^T} - b_X \right) + A_B\vec{\delta B} + A_{C^T}\vec{\delta C^T} + A_t\delta t
\end{multline}
where we now define
\begin{equation}
    r_{\mathrm{pOSE}} = A_X\vec{Rg(B_0C_0^T)+t_0\mathbbm{1}^T} - b_X
\end{equation}
with
\begin{equation}
    A_B = A_X(I \otimes R)\Gamma_g(C_0 \otimes I), \quad  A_{C^T} = A_X(I \otimes R)\Gamma_g(I \otimes B_0),\quad A_t = A_X(\mathbbm{1} \otimes I),
\end{equation}
where $\Gamma_g$ maps the elements from $BC^T$ to $g(BC^T)$ such that
\begin{equation}
    \vec{g(BC^T)} = \Gamma_g\vec{BC^T}.
\end{equation}
Since the weights $a$ are applied to the singular values of $BC^T$, the weighted nuclear norm terms can be written as \eqref{eq:sup_regB} and \eqref{eq:sup_regC}, similarly to the Low-rank Matrix Recovery problem. The residuals and jacobian in Algorithm 1 for the Non-rigid Strucuture Recovery problem can be computed as
\begin{equation}
    J = \begin{bmatrix} A_B & A_{C^T} & A_t\\
    A_{\mathrm{regB}} & \mathbf{0} & \mathbf{0}\\
    \mathbf{0} & A_{\mathrm{regC}} & \mathbf{0}
    \end{bmatrix}, \quad r_a = \A_a(B_0C_0^T) + b_a = \begin{bmatrix}
    r_{\mathrm{pOSE}}\\r_{\mathrm{regB}}\\r_{\mathrm{regC}}
    \end{bmatrix},
\end{equation}
and the translation is also added to the auxiliary variable $z$ in Algorithm 1, i.e., $z = [\vec{B};\vec{C^T};t]$.

\section{Results on Back, Heart and Paper Datasets}
\begin{figure}[t]
\centering
\begin{subfigure}{1\textwidth}
\includegraphics[height=1.745cm]{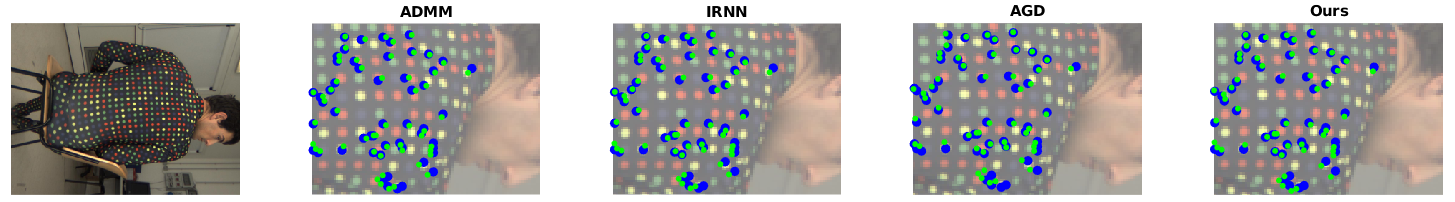}
\end{subfigure}
\begin{subfigure}{1\textwidth}
\includegraphics[height=1.85cm]{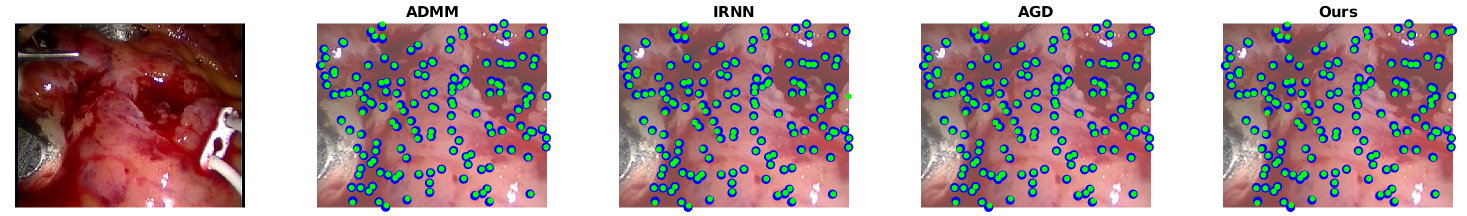}
\end{subfigure}
\begin{subfigure}{1\textwidth}
\includegraphics[height=1.83cm]{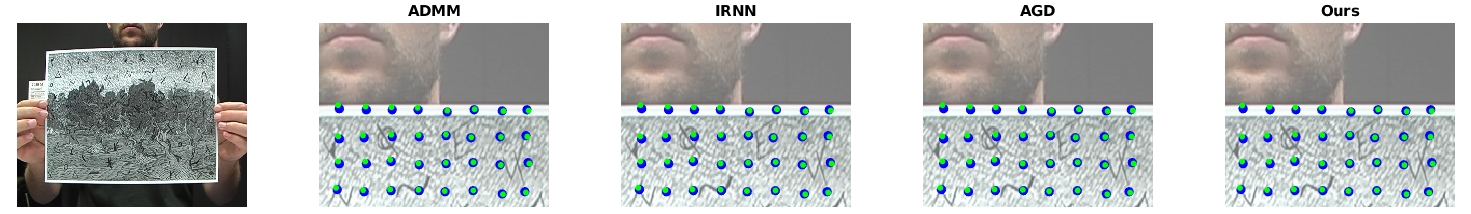}
\end{subfigure}
\caption{Comparison between reprojections (green) and 2D measurements (blue) obtained in Section 5.2 for Back (top), Heart (middle), and Paper (bottow) datasets. }
\setlength{\belowcaptionskip}{-40pt}
\label{fig:back_heart_paper_results}
\end{figure}

In Figure~\ref{fig:back_heart_paper_results} we show an example of the reprojection errors obtained for the Back, Heart and Paper datasets in Section 5.2, for the weighted nuclear norm regularization and $\eta = 0.05$ (near perspective). Even though the qualitative difference between the methods is not visible (note the y-axis scale on the plots in Figure 1), the second-order method was still able to obtain a lower loss than all the first-order methods. 

\section{Results on NRSfM Challange Datasets}
In this section we provide all results obtained with the weighted nuclear norm for the perspective camera model of the NRSfM Challange datasets. These include the the log-losses (Table~\ref{tab:log_loss}) and 3D reconstruction errors (Table~\ref{tab:rec_err}) for the ADMM and our method, in each of the six sequences (Circle, Flyby, Line, Semi-circle, Tricky, Zigzag) of the five datasets. Recall that the values in Tables 1 and 2 in the main text correspond to the average over the six sequences, for each dataset. In Figures~\ref{fig:nrsfm_1} and \ref{fig:nrsfm_2} we also show the qualitative comparison between the 3D reconstruction obtained with the two methods and the provided 3D ground-truth structure, for each sequence. 

Note that our method is always able to obtain a lower loss compared to the ADMM, and the 3D reconstruction is always as good or much better (see the cases of Ballon-Semi-circle, Balloon-Tricky, Paper-Tricky, and Stretch-Flyby). The only exception was the sequence Tearing-Zigzag, where a lower loss actually resulted in a worse 3D reconstruction, which might be explained by incorrect modeling ($K = 2$ might be too low for this sequence).  

\begin{table}[t]
\setlength{\abovecaptionskip}{-1pt}
\caption{Log-loss on each for all sequences of the perspective datasets.}
\begin{tabular}{>{\centering}p{0.12\textwidth}>{\centering}p{0.2\textwidth}>{\centering}p{0.10\textwidth}>{\centering}p{0.10\textwidth}>{\centering}p{0.10\textwidth}>{\centering}p{0.10\textwidth}>{\centering}p{0.10\textwidth}>{\centering\arraybackslash}p{0.10\textwidth}}
  & \multicolumn{1}{c|}{Method \textbackslash Sequence}  & Circle & Flyby & Line & Semi-circle & Tricky & Zigzag \\ \cline{2-8} 
\multicolumn{1}{c}{\multirow{2}{*}{Articulated}} & \multicolumn{1}{c|}{ADMM-WNN} & -1.822 & -2.849 & -3.797 & -3.405 &-3.009  & -2.517 \\
\multicolumn{1}{c}{} & \multicolumn{1}{c|}{Ours-WNN} & \textbf{-1.825} & \textbf{-2.853} & \textbf{-3.845} & \textbf{-3.408} & \textbf{-3.030} & \textbf{-2.753} \\ \hline \hline
\multirow{2}{*}{Balloon} & \multicolumn{1}{c|}{ADMM-WNN} & -2.232 & -2.977 & -3.130 & -2.607 &-2.380  &  -3.834\\
 & \multicolumn{1}{c|}{Ours-WNN}& \textbf{-2.465} & \textbf{-3.325} & \textbf{-3.096} & \textbf{-2.949} & \textbf{-2.934}  & \textbf{-4.037}\\\hline \hline \multirow{2}{*}{Paper} & \multicolumn{1}{c|}{ADMM-WNN} & -1.451 & -3.037 & -3.822 & -3.171  &-3.112  & -3.473 \\ 
 & \multicolumn{1}{c|}{Ours-WNN} & \textbf{-2.107 } & \textbf{-3.037} & \textbf{-3.823} & \textbf{-3.171} & \textbf{-3.809} & \textbf{-3.498}\\\hline \hline
 \multirow{2}{*}{Stretch} & \multicolumn{1}{c|}{ADMM-WNN} & -2.267 & -2.253 & -3.629 & -2.722 & -3.574 & -4.542 \\
 & \multicolumn{1}{c|}{Ours-WNN} &\textbf{-2.275}  & \textbf{-3.153} & \textbf{-3.846} & \textbf{-2.724} & \textbf{-3.578} &\textbf{-4.546}\\\hline \hline
 \multirow{2}{*}{Tearing} & \multicolumn{1}{c|}{ADMM-WNN} & -1.834 & -1.154 & -3.302 & -1.888 & -3.504  & -1.612 \\
 & \multicolumn{1}{c|}{Ours-WNN} & \textbf{-2.184} & \textbf{-1.662} & \textbf{-3.302} & \textbf{-2.067} & \textbf{-3.521} & \textbf{-2.017}
\end{tabular}
\label{tab:log_loss}
\end{table}
\begin{table}[h]
\setlength{\abovecaptionskip}{-1pt}
\caption{3D reconstruction error, in millimeters, on each for all sequences of the perspective datasets  relatively to the provided ground-truth structure.}
\begin{tabular}{>{\centering}p{0.12\textwidth}>{\centering}p{0.2\textwidth}>{\centering}p{0.10\textwidth}>{\centering}p{0.10\textwidth}>{\centering}p{0.10\textwidth}>{\centering}p{0.10\textwidth}>{\centering}p{0.10\textwidth}>{\centering\arraybackslash}p{0.10\textwidth}}
  & \multicolumn{1}{c|}{Method \textbackslash Sequence}  & Circle & Flyby & Line & Semi-circle & Tricky & Zigzag \\ \cline{2-8} 
\multicolumn{1}{c}{\multirow{2}{*}{Articulated}} & \multicolumn{1}{c|}{ADMM-WNN} & 15.69 & \textbf{9.52} & 13.33 & 16.49 & \textbf{27.77} & 26.65 \\
\multicolumn{1}{c}{} & \multicolumn{1}{c|}{Ours-WNN} & \textbf{13.84} &  9.67 & \textbf{12.35}& \textbf{14.32} & 32.49 & \textbf{16.52}\\ \hline \hline
\multirow{2}{*}{Balloon} & \multicolumn{1}{c|}{ADMM-WNN} & 3.56 & \textbf{2.64} & \textbf{4.73}  & 16.06 & 24.46& 2.23 \\
 & \multicolumn{1}{c|}{Ours-WNN}& \textbf{2.07}  & 2.92 & 4.78 &\textbf{ 5.48} & \textbf{20.19}  & \textbf{2.19} \\\hline \hline \multirow{2}{*}{Paper} & \multicolumn{1}{c|}{ADMM-WNN} & 8.62 & \textbf{4.71} & \textbf{6.71} & 6.12 & 30.45 & \textbf{4.22} \\ 
 & \multicolumn{1}{c|}{Ours-WNN} & \textbf{1.98} & 4.75 & 7.06  & \textbf{6.02} & \textbf{9.80} & 4.45 \\\hline \hline
 \multirow{2}{*}{Stretch} & \multicolumn{1}{c|}{ADMM-WNN} & \textbf{2.59} & 16.86 & \textbf{4.78} & 5.68 & 15.65 & \textbf{2.80}  \\
 & \multicolumn{1}{c|}{Ours-WNN} & 2.69 & \textbf{2.85} & 6.74  & \textbf{5.58} & \textbf{14.93} & 2.84 \\\hline \hline
 \multirow{2}{*}{Tearing} & \multicolumn{1}{c|}{ADMM-WNN} & 5.10 & 10.94 & \textbf{8.93} & 5.05 &  18.57 & \textbf{7.09} \\
 & \multicolumn{1}{c|}{Ours-WNN} & \textbf{4.25}  & \textbf{7.15} &  9.17 & \textbf{4.87}& \textbf{16.98}  & 8.12
\end{tabular}
\label{tab:rec_err}
\end{table}

\newpage
\iffalse
\begin{figure}
\begin{subfigure}{0.01\textwidth}
\rotatebox[origin=t]{90}{Circle}
\end{subfigure}
\begin{subfigure}{0.19\textwidth}
\includegraphics[height=2cm]{images/articulated_circle_3D.png}
\end{subfigure}
\begin{subfigure}{0.19\textwidth}
\includegraphics[height=2cm]{images/articulated_flyby_3D.png}
\end{subfigure}
\begin{subfigure}{0.19\textwidth}
\includegraphics[height=2cm]{images/articulated_line_3D.png}
\end{subfigure}

\begin{subfigure}{0.19\textwidth}
\includegraphics[height=2cm]{images/articulated_semi_circle_3D.png}
\end{subfigure}
\begin{subfigure}{0.19\textwidth}
\includegraphics[height=2cm]{images/articulated_tricky_3D.png}
\end{subfigure}
\begin{subfigure}{0.19\textwidth}
\includegraphics[height=2cm]{images/articulated_zigzag_3D.png}
\end{subfigure}
\end{figure}
\fi

\begin{sidewaysfigure}
\begin{subfigure}{1\textwidth}
\hspace{1.5cm}Circle Sequence 
\hspace{5.5cm}Flyby Sequence
\hspace{5cm}Line Sequence
\end{subfigure}
\vspace{0.1cm}

\begin{subfigure}{1\textwidth}
\includegraphics[height=2.25cm]{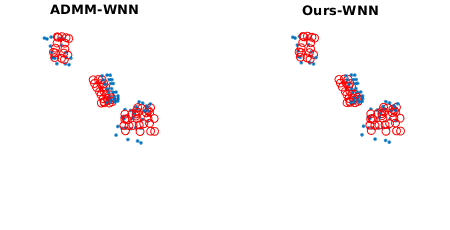}
\hspace{3cm}
\includegraphics[height=2.25cm]{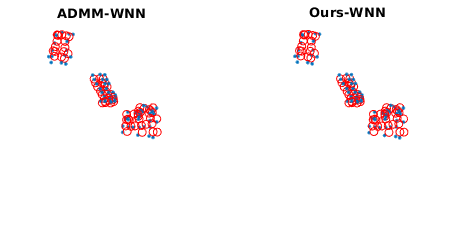}
\hspace{2.5cm}
\includegraphics[height=2.25cm]{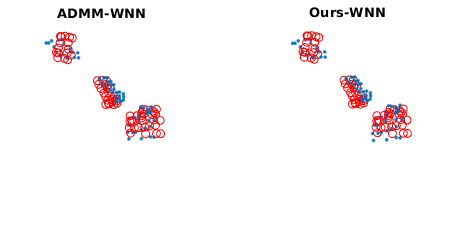}
\end{subfigure}

\begin{subfigure}{1\textwidth}
\includegraphics[height=2.25cm]{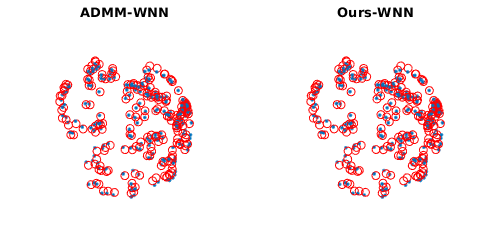}
\hspace{2.5cm}
\includegraphics[height=2.25cm]{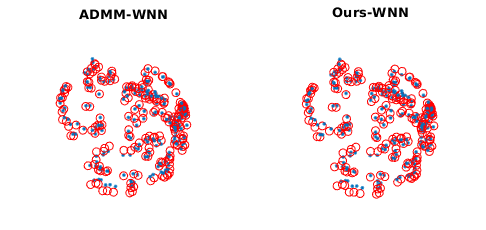}
\hspace{2.15cm}
\includegraphics[height=2.25cm]{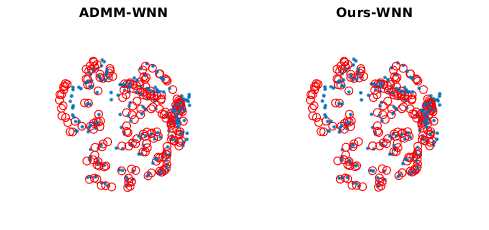}
\end{subfigure}

\begin{subfigure}{1\textwidth}
\includegraphics[height=2.25cm]{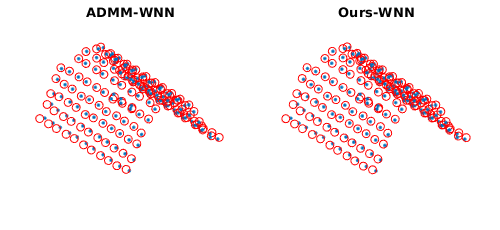}
\hspace{2.5cm}
\includegraphics[height=2.25cm]{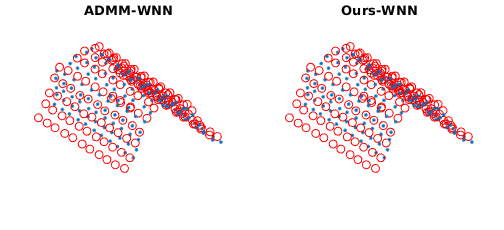}
\hspace{2cm}
\includegraphics[height=2.25cm]{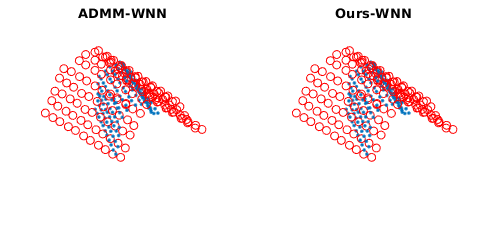}
\end{subfigure}

\begin{subfigure}{1\textwidth}
\includegraphics[height=2.25cm]{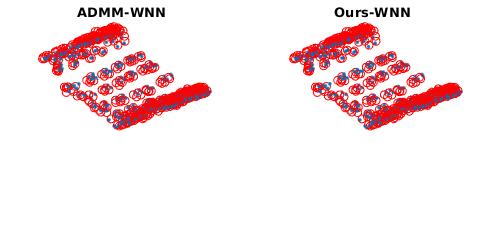}
\hspace{2.7cm}
\includegraphics[height=2.25cm]{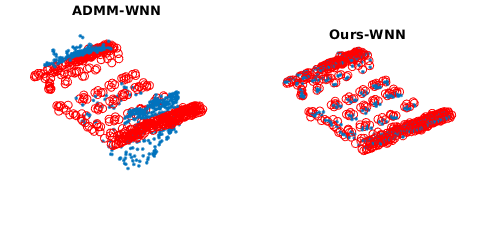}
\hspace{2cm}
\includegraphics[height=2.25cm]{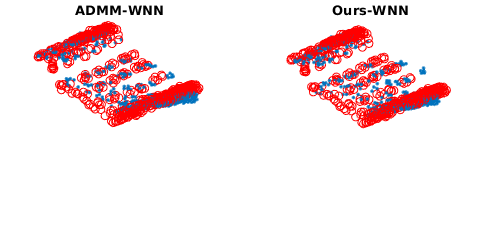}
\end{subfigure}

\begin{subfigure}{1\textwidth}
\includegraphics[height=2.25cm]{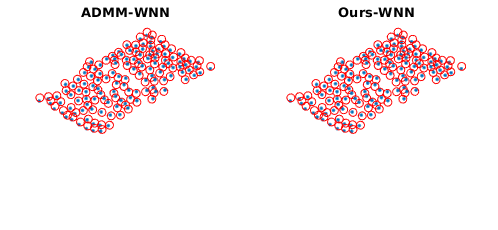}
\hspace{2.7cm}
\includegraphics[height=2.25cm]{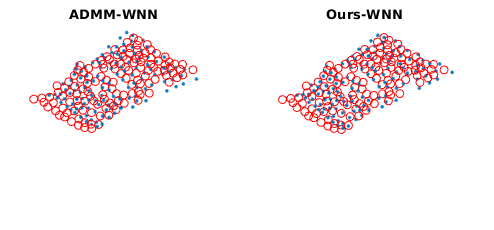}
\hspace{2cm}
\includegraphics[height=2.25cm]{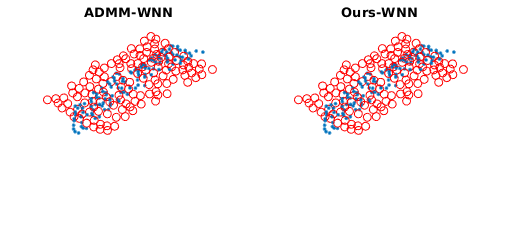}
\end{subfigure}
\caption{Comparison between the estimated (blue) and provided (red) 3D structure for the sequences Circle, Flyby and Line.}
\label{fig:nrsfm_1}
\end{sidewaysfigure}

\begin{sidewaysfigure}
\begin{subfigure}{1\textwidth}
\hspace{1cm}Semi-circle Sequence 
\hspace{5.25cm}Tricky Sequence
\hspace{5cm}Zigzag Sequence
\end{subfigure}
\vspace{0.1cm}

\begin{subfigure}{1\textwidth}
\includegraphics[height=2.25cm]{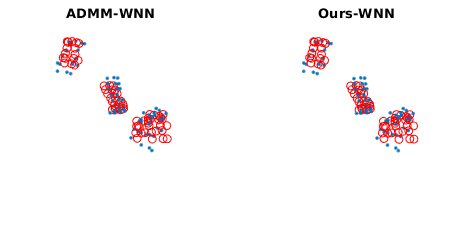}
\hspace{3cm}
\includegraphics[height=2.25cm]{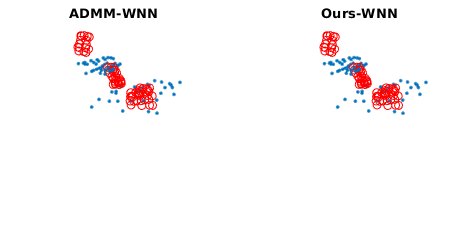}
\hspace{2.5cm}
\includegraphics[height=2.25cm]{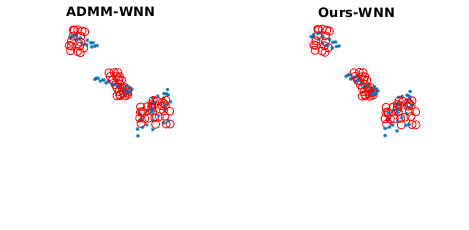}
\end{subfigure}

\begin{subfigure}{1\textwidth}
\includegraphics[height=2.25cm]{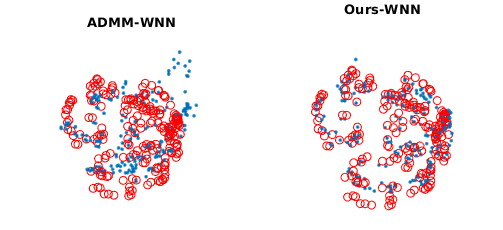}
\hspace{2.5cm}
\includegraphics[height=2.25cm]{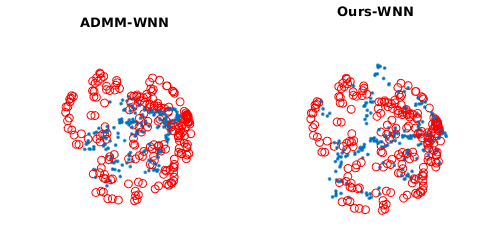}
\hspace{2.15cm}
\includegraphics[height=2.25cm]{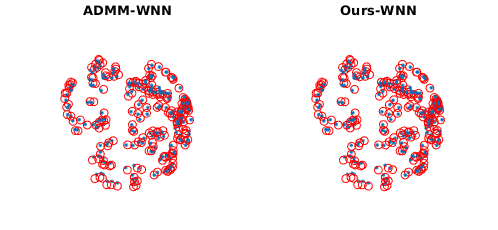}
\end{subfigure}

\begin{subfigure}{1\textwidth}
\includegraphics[height=2.25cm]{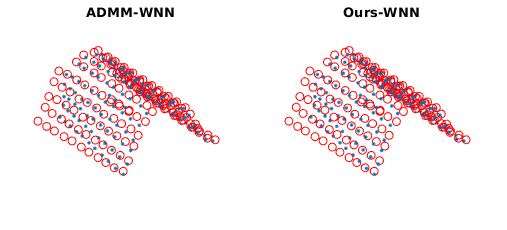}
\hspace{2.5cm}
\includegraphics[height=2.25cm]{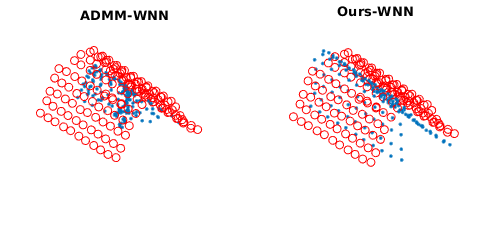}
\hspace{2cm}
\includegraphics[height=2.25cm]{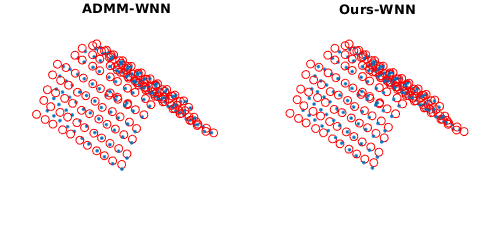}
\end{subfigure}

\begin{subfigure}{1\textwidth}
\includegraphics[height=2.25cm]{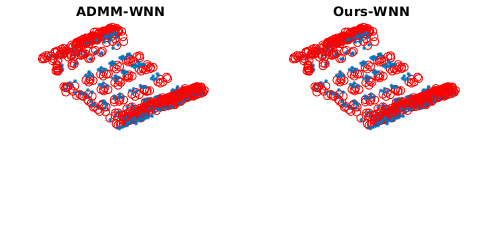}
\hspace{2.7cm}
\includegraphics[height=2.25cm]{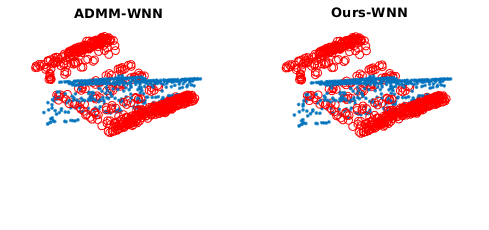}
\hspace{2cm}
\includegraphics[height=2.25cm]{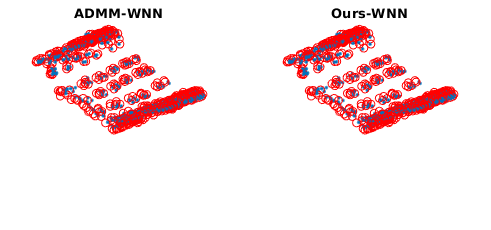}
\end{subfigure}

\begin{subfigure}{1\textwidth}
\includegraphics[height=2.25cm]{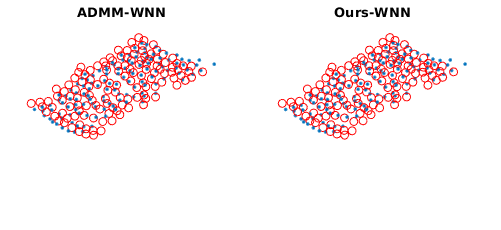}
\hspace{2.7cm}
\includegraphics[height=2.25cm]{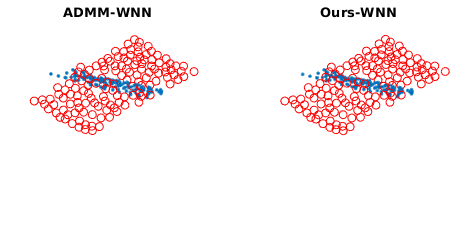}
\hspace{2cm}
\includegraphics[height=2.25cm]{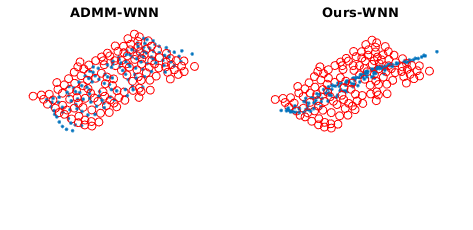}
\end{subfigure}
\caption{Comparison between the estimated (blue) and provided (red) 3D structure for the sequences Semi-circle, Tricky and Zigzag.}
\label{fig:nrsfm_2}
\end{sidewaysfigure}
%\end{document}

%{\small
%\bibliographystyle{plain}
%\bibliography{newlib}
%}

\end{document}